\documentclass{article}
\usepackage{ijcai25}

\usepackage{times}
\usepackage{soul}
\usepackage{url}
\usepackage[hidelinks]{hyperref}
\usepackage[utf8]{inputenc}
\usepackage[small]{caption}
\usepackage{graphicx}
\usepackage{amsmath}
\usepackage{amsthm}
\usepackage{booktabs}
\usepackage{algorithm}
\usepackage{algorithmic}
\usepackage[switch]{lineno}

\usepackage{subcaption}
\usepackage{amsfonts}
\usepackage{multicol}
\usepackage{makecell}
\usepackage{multirow}
\usepackage{xcolor}
\newcommand{\bc}[1]{{#1}}
\title{LiBOG: Lifelong Learning for Black-Box Optimizer Generation}

\author{
Jiyuan Pei$^1$
\and
Yi Mei$^{1}$\footnote{Yi Mei is the corresponding author.}\and
Jialin Liu$^2$\And
Mengjie Zhang$^1$\\
\affiliations
$^1$Victoria University of Wellington\\
$^2$Lingnan University\\
\emails
jiyuan.pei@vuw.ac.nz,
yi.mei@ecs.vuw.ac.nz,
jialin.liu@ln.edu.hk,
mengjie.zhang@ecs.vuw.ac.nz
}
\begin{document}

\maketitle

\begin{abstract}

% Meta-Black-Box Optimization (MetaBBO) garners attention due to its success in automating the configuration and generation of black-box optimizers, significantly reducing the human effort required for optimizer design and discovering optimizers with higher performance than classic human-designed optimizers. However, existing MetaBBO methods conduct one-off training under the assumption that extensive and representative training problems are pre-available. This assumption is often impractical in real-world scenarios, where diverse and evolving problems with shifting characteristics continually arise. Consequently, there is a pressing need for methods that can continuously learn from new problems encountered on-the-fly and progressively enhance their capabilities.
% In this work, we explore the novel paradigm of lifelong learning in MetaBBO and introduce LiBOG, a novel approach designed to learn from sequentially encountered problems and generate high-performance BBO optimizers. LiBOG consolidates knowledge both across tasks and within tasks to mitigate catastrophic forgetting.
% Extensive experiments under various lifelong learning scenarios demonstrate LiBOGs effectiveness in learning to generate high-performance optimizers in a lifelong learning manner, addressing catastrophic forgetting with impressive stability while maintaining strong plasticity to learn new tasks.
% % Furthermore, LiBOG exhibits the ability to learn generalized knowledge across sequential tasks and perform backward transfer.

Meta-Black-Box Optimization (MetaBBO) garners attention due to its success in automating the configuration and generation of black-box optimizers, significantly reducing the human effort required for optimizer design and discovering optimizers with higher performance than classic human-designed optimizers. However, existing MetaBBO methods conduct one-off training under the assumption that a stationary problem distribution with extensive and representative training problem samples is pre-available. This assumption is often impractical in real-world scenarios, where diverse problems following shifting distribution continually arise. Consequently, there is a pressing need for methods that can continuously learn from new problems encountered on-the-fly and progressively enhance their capabilities.
In this work, we explore a novel paradigm of lifelong learning in MetaBBO and introduce LiBOG, a novel approach designed to learn from sequentially encountered problems and generate high-performance optimizers for Black-Box Optimization (BBO). LiBOG consolidates knowledge both across tasks and within tasks to mitigate catastrophic forgetting.
Extensive experiments demonstrate LiBOG's effectiveness in learning to generate high-performance optimizers in a lifelong learning manner, addressing catastrophic forgetting while maintaining plasticity to learn new tasks.

\end{abstract}

\section{Introduction}

\begin{figure}[htbp]
\centering
% 第一个子图
\begin{subfigure}[b]{0.48\linewidth}
\centering
\includegraphics[width=\linewidth]{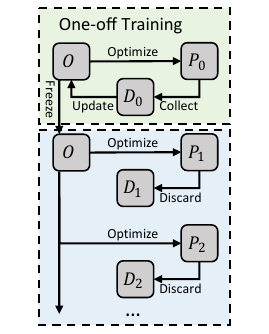} 
\caption{MetaBBO}
\label{fig:MetaBBO_traditional}
\end{subfigure}
% \hfill 
\begin{subfigure}[b]{0.48\linewidth}
\centering
\includegraphics[width=\linewidth]{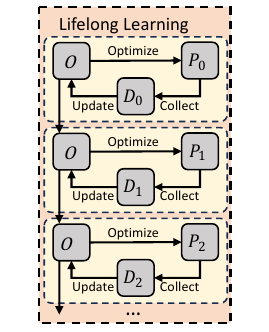}
\caption{Lifelong Learning MetaBBO}
\label{fig:MetaBBO_lifelong}
\end{subfigure}
\caption{Existing MetaBBO methods train the BBO optimizer $O$ with data $D_0$ obtained from the problem distribution $P_0$ available during the training phase, then freeze the model after training to solve the newly arising problems from distributions $(P_1,P_2,\dots)$. New data obtained from subsequent problems are discarded. In contrast, lifelong learning MetaBBO utilizes the data obtained from each encountered problem to update the optimizer continually.}
\label{fig:MetaBBOs}
\end{figure}

Black-Box Optimization (BBO) solves optimization problems by using only the output of the objective function, without requiring knowledge of its internal structure. It is commonly used in tasks with complex structures like metal manufacturing~\cite{bbometal}, protein docking~\cite{Tsaban_2022}, and hyper-parameter tuning of learning algorithms~\cite{pyhopper,gu2021optimizing}. 
% BBO optimizers iteratively evaluate solutions and generate new ones based on human-designed update rules. 
Typically, designing and tuning BBO optimizers is labor-intensive and requires expertise.
Meta-Black-Box Optimization (MetaBBO) automates this process with machine learning, greatly reducing manual effort. By training on \bc{BBO}
% a set of training problems
problems sampled from a given distribution, MetaBBO learns to directly propose solutions~\cite{chen2017ICML,TV2019meta} \bc{to some given problems}, configure optimizers with expert-derived solution update rules~\cite{DE-DDQN,GSF-DQN,Lu2020A,chaybouti2022metalearning}, or generate optimizers by automatically generating solution update rules~\cite{chen2024symbol}. Existing MetaBBO methods rely on one-off training, where the optimizer is trained on
% a pre-defined fixed problem set, 
problems sampled from a pre-defined fixed distribution,
and then is directly applied to test problems (e.g., learned parameters are frozen)~\cite{Lu2020A,surveyL20}, as shown in Fig~\ref{fig:MetaBBO_traditional}.

In real-world optimization scenarios, the distribution of optimization problems commonly varies over time, leading to the emergence of new problems with different but related characteristics.
For example, the scheduling problem faced by a manufacturer may vary from season to season due to dynamic factors such as demands and resource supplies. 
% For example, energy scheduling of a new time period can change significantly due to sudden events such as wars or price fluctuations in natural resources, and hyper-parameter tuning of new learning algorithms may change substantially with the new model structures, new data sets, and new learning tasks.
% and the characteristics of structural design problems may shift substantially with the development of new materials.
Directly applying the pre-trained optimizer to solve new problems 
% is usually
could be
ineffective~\cite{howgoodNCO,online-offline-aos,surveyL20,yang2025graph}.
% due to the limited generalization ability.
% Redesigning the optimizer is usually required when facing new problems. 
To address this issue, the traditional way in practice typically involves modifications of the optimizer by human experts when problem characteristics change significantly~\cite{hoos2012automated}.
% to improve performance on new problems without sacrificing performance on previously encountered problems. 
However, existing MetaBBO methods encounter limitations in this adaptability. MetaBBO approaches capable of continuously learning from emerging problems to enhance their ability in a lifelong learning manner, as shown in Fig~\ref{fig:MetaBBO_lifelong}, are desired.

% \begin{figure}
% \centering
% \includegraphics[width=.9\linewidth]{MetaBBO.pdf}
% \caption{Existing MetaBBO methods (left) train the BBO optimizer $O$ with data $D_0$ obtained from the problem distribution $P_0$ available during the training phase, then freeze the model after training to solve the subsequent problem distribution $(P_1,P_2,\dots)$. Data obtained from subsequent problems are discarded. In contrast, lifelong learning MetaBBO (right) utilizes the data obtained from each encountered problem set to update the optimizer continually. }
% \label{fig:MetaBBOs}
% \end{figure}

Lifelong learning~\cite{surveyCL} is a natural paradigm to continually learn/adapt BBO solvers when facing changing problem distribution over time. Although lifelong learning has achieved great success in common machine learning tasks~\cite{ewc,wang2020incrementalTNNLS}, to the best of our knowledge, no study has investigated it in the context of MetaBBO.
% Lifelong learning refers to the paradigm in which models learn to expert multiple tasks that are available sequentially~\cite{surveyCL}.
% rather than all at once~\cite{}.
% Each task is represented by a identical distribution of data. The key challenge in lifelong learning is to achieve good plasticity of learn the knowledge of new tasks efficiently while maintain good stability of preserving learned knowledge about the trained tasks. The phenomenon of reducing performance on previously trained data after learning from new data with different distributions is known as catastrophic forgetting. Mostly studied, catastrophic forgetting occurs between learning tasks as the data from tasks experience significant differences. While in reinforcement learning field where the training data is obtained by the interaction between the learner model and the given environment, catastrophic forgetting may also occurs within the learning process of a single task.
In particular, a major challenge in lifelong learning is catastrophic forgetting~\cite{surveyCRL,surveyCL}, i.e., models' performance on previous data distributions reduces significantly after training on a new distribution.  Mostly studied, catastrophic forgetting usually occurs between tasks as the data from different tasks exhibit significant differences. However, catastrophic forgetting may also occur 
% within the learning process of 
when learning a single task, especially in reinforcement learning (RL)~\cite{igl2021transient,lan2023memoryefficient,zhang2023catastrophic}.

In this paper, we focus on the unexplored paradigm of lifelong learning MetaBBO, with the aim of training a single model capable of generating high-performance optimizers for any problem drawn from previously learned distributions,
in scenarios where problem distributions arrive sequentially over time.
To achieve it, we present \textbf{LiBOG}, a novel approach with \underline{\textbf{li}}felong learning for \underline{\textbf{b}}lack-box \underline{\textbf{o}}ptimizer \underline{\textbf{g}}eneration. LiBOG takes the problems following the same distribution as a task and learns from a sequence of different tasks with lifelong RL.
LiBOG features an inter-task consolidation process to preserve the learned knowledge of previous tasks. 
% Besides, our investigation revealed that catastrophic forgetting may also occur within one single task and significantly impacts learning performance in the context of MetaBBO. 
We further propose a novel intra-task consolidation method, named elite behavior consolidation, to address forgetting in one single task which could significantly impact learning performance in the context of MetaBBO,
% While preserving the learned knowledge, the intra-task consolidation method also contributes to maintaining plasticity, improving the learning effectiveness on the new task. 
% Fig~\ref{fig:overall_process} demonstrates the overall process of LiBOG.

We verify the effectiveness of LiBOG in generating high-performance optimizers for solving problems from all learned distributions through extensive experimental studies. The results demonstrate that LiBOG not only significantly mitigates catastrophic forgetting but also exhibits the capability to learn general knowledge from sequentially arriving problem distributions. Further sensitivity analysis and ablation study verify the stability of LiBOG and the contribution of each component to LiBOG's overall performance.
% , achieving backward knowledge transfer in certain cases. 
% Furthermore, LiBOG outperforms the SOTA human-designed optimizer and SOTA MetaBBO method that is fine-tuned on sequentially arrived problem distributions.
% embedded with widely used lifelong learning approaches,
% , like fine-tuning and elastic weight consolidation (EWC)~\cite{ewc}.
% , one of the most widely used lifelong learning methods, and
% demonstrates superior problem-solving ability compared to single-task training, where the model is trained solely on one identical task in isolation, without the effect of catastrophic forgetting due to the difference between tasks.

Our main contributions are as follows. (i) We propose the paradigm of lifelong learning for BBO optimizer generation, which is the first study in this area, to the best of our knowledge. 
% (ii) We empirically investigate the catastrophic forgetting in the SOTA MetaBBO methods for learning BBO optimizer generation~\cite{} and reveal the significant impact of catastrophic forgetting occurring both between tasks and within one single task.
(ii) We present elite behavior consolidation, a novel intra-task consolidation approach to address catastrophic forgetting and improve plasticity, and based on it, LiBOG, a novel method of lifelong learning to BBO optimizer generation.
(iii) We verify the effectiveness of the proposed methods with extensive experimental studies, together with detailed analysis and in-depth discussion.

% \begin{figure*}
% \centering
% \includegraphics[width=0.5\linewidth]{}
% \caption{Caption}
% \label{fig:overall_process}
% \end{figure*}

\section{Background}

\subsection{MetaBBO}

% Black-Box Optimization (BBO) is the framework for solving optimization problems in which the objective function is treated as a "black box," with no assumptions or access to its internal structure or analytical properties. Instead, BBO optimizers rely exclusively on the output of the objective function to guide the optimization process. 
% BBO optimizers own the flexibility to be applied to problems with non-differentiable, noisy, or even discontinuous objective functions. These attributes have made BBO a valuable tool in addressing complex, real-world optimization challenges. Examples of its applications include hyperparameter tuning in machine learning algorithms, optimizing energy scheduling systems, structural design processes, and drug discovery, where objective functions are often computationally expensive or lack explicit formulations.
% Prominent examples of BBO techniques include evolutionary algorithms, Bayesian optimization, and simulated annealing, all of which have demonstrated effectiveness across diverse application domains.
Black-box optimization relies exclusively on the output of the objective function to guide the optimization process~\cite{Audet_2016}. 
A typical BBO optimizer operates iteratively, wherein incumbent solutions are evaluated based on their objective values, and modified by solution updating rules with the aim of achieving superior performance~\cite{kashif2018metaheuristic}. 
Examples of BBO optimizers include 
evolution strategy~\cite{Beyer_2002},
differential evolution (DE)~\cite{Storn_1997},
Bayesian optimization~\cite{BOsurvey}, and simulated annealing~\cite{10.1126/science.220.4598.671}. 
% The iterative evaluation-and-generation mechanism allows BBO to address a wide range of complex optimization problems without requiring gradient information or detailed insights into the underlying problem.
Typically, BBO optimizers are characterized by a wide array of tunable parameters, 
% such as population size, mutation rates, and exploration-exploitation strategies, 
which significantly influence their performance. The design and configuration of these parameters are often computationally expensive and labor-intensive, requiring substantial expertise and iterative experimentation.

Meta-Black-Box Optimization (MetaBBO) is an emerging framework that leverages machine learning techniques to automate the design of black-box optimizers, significantly reducing the reliance on manual expertise and effort~\cite{chen2017ICML,gomes2021meta,NEURIPS2023_232eee8e}. 
% By learning from data, MetaBBO provides a systematic way to improve optimization performance while minimizing human intervention.
% MetaBBO encompasses three primary categories to conduct data-driven optimizer design. The first category of 
Some MetaBBO methods focus on training end-to-end models to directly generate new solutions~\cite{chen2017ICML,TV2019meta}. Despite their success, end-to-end methods often suffer from poor generalization and limited interpretability~\cite{howgoodNCO}.
% The second category is
Some other methods learn to configure human-designed optimizers, including parameter tuning and selecting solution updating rules~\cite{DE-DDQN,GSF-DQN,Lu2020A,chaybouti2022metalearning}. However, the corresponding methods are inherently limited by the dependence of pre-existing human-crafted solution updating rules.
% The third category is learning to construct updating rules, replacing human-designed ones. 
In contrast, the recent study~\cite{chen2024symbol} leverages symbolic equation learning and deep reinforcement learning to construct updating rules and outperforms SOTA expert-designed methods with human-crafted rules.
Following the conventional machine learning paradigm, existing MetaBBO methods are typically trained on 
% a given set of training problems or samples of a pre-defined problem distribution,
problems sampled from a pre-defined distribution,
with the optimizer's parameters fixed upon completion of training. These trained models are then evaluated on separate test sets to assess their performance.
To achieve good performance across a broader range of problems, current MetaBBO approaches typically involve a large number of diverse problems in the one-off training. 

\subsection{Lifelong Learning}

% Lifelong learning, as known as continual learning, is a paradigm in which models learn to master a sequence of tasks presented sequentially rather than simultaneously~\cite{}, where each task is characterized by an identical data distribution.

% While traditional machine learning aims at learning to expert one single task, lifelong learning, also known as continual learning, requires learners to learn continuously over time from many different tasks sequentially to progressively enhance their capabilities on the tasks, just like the way that humans learn in their whole life~\cite{Thrun1998lifelong}.

% While conventional machine learning focuses on learning to expert one single task, l
Lifelong learning, also known as continual learning, requires learners to learn a sequence of different tasks to progressively enhance their capabilities on the tasks, just like the way that humans learn in their whole life~\cite{Thrun1998lifelong}. 
In lifelong learning, each task is represented by a distinct data distribution~\cite{surveyCL}. Lifelong learning primarily aims to efficiently train on the data distribution of the current task while minimizing reliance on data from the distributions of previously encountered tasks. 
The key challenge in lifelong learning is to balance plasticity, the ability to efficiently acquire knowledge from new tasks, with stability, the capacity to preserve learned knowledge of previous tasks. The degradation of performance on learned tasks after training on new data with different distributions is known as \textit{catastrophic forgetting}. This phenomenon is predominantly observed between tasks due to significant differences in their data.

% Reinforcement learning (RL) requires the agent to learn to make decisions under a specific environment by interacting with the environment and learning from the obtained experiences~\cite{Kaelbling1996reinforcement}. An environment is typically described by an MDP.
Lifelong learning in the RL context refers to the problem of how an agent learns from a series of different environments to make good decisions on each of them~\cite{Abel2018policy,surveyCRL}. Each environment, \bc{commonly defined by} a Markov decision process (MDP), corresponds to a task.
Specifically in RL, catastrophic forgetting can occur not only between tasks but also during the learning of a single task~\cite{igl2021transient}. Training data, i.e., experiences, are typically generated iteratively through interactions between the policy model and the given environment. These experiences exhibit temporal correlations, and the distribution of the collected experiences changes with the policy updating,
% and its behavior pattern change, 
contributing to catastrophic forgetting~\cite{igl2021transient}. % RL problems with higher instability of experience distribution within one single task is more like impacted by the catastrophic forgetting。
Typically, it is addressed by extensive resampling and large experience memory, significantly increasing the storage requirements and computational costs~\cite{lan2023memoryefficient}.

% The formulation of lifelong RL can be divided two parts~\cite{Abel2018policy}: (i) the task-independent state space $\mathcal{S}$ and action space $\mathcal{A}$, and (ii) the task-specific reward function $\mathcal{R}$, transition function $\mathcal{P}$, horizon of action steps $\mathcal{H}$, and initial state probability distribution $\rho$, following a fixed but unknown distribution $\mathcal{D}$. In the lifelong RL problem, each task is to make decisions in the environment represented by MDP ($(\mathcal{S},\mathcal{A})\cup(\mathcal{R},\mathcal{P},\mathcal{H},\rho)$).

% The more different elements there are between tasks, the less applicable experiences obtained from previous tasks are to the current task, making the lifelong learning scenario more challenging. Most existing lifelong RL studies focus on only changing $\mathcal{R}$~\cite{Abel2018policy,wang2019incrementalTM,tasse2020logicalICML,tasse2022generalisationICLR} or $\mathcal{R} \cup \mathcal{P}$~\cite{wang2020incrementalTNNLS,Lecarpentier2021lipschitzAAAI,wang2021lifelongTNNLS,deng2023incrementalTAI,xie2022lifelongPMLR,mendez2022modularICML}, leaving the rest part of MDP fixed between tasks.

% \section{Related Work}
% \subsection{Learn to Generate BBO Optimizer}
% \subsection{Catastrophic Forgetting in Lifelong Reinforcement Learning}
% \subsection{Lifelong Learning with Consolidation}

% \section{Intra-Task Catastrophic Forgetting}

\section{LiBOG}

\begin{figure*}[htbp]
\centering
\includegraphics[width=\linewidth]{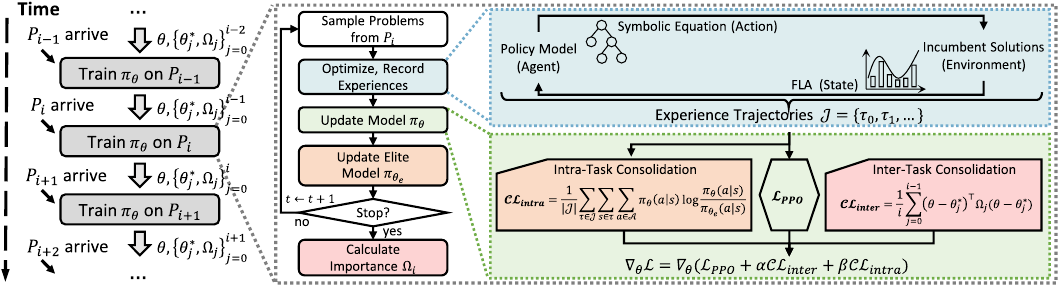}
\caption{Illustration of LiBOG, with different problem distributions $P_{i-1},P_{i},\dots$ sequentially arrive.}
\label{fig:LiBOG}
\end{figure*}

A typical BBO optimizer improves the objective value by iteratively modifying incumbent solutions with solution updating rules, inherently forming a sequential decision-making process~\cite{handoko2014RL,chaybouti2022metalearning}.
Solution updating rules
% responsible for improving the objective value of solutions 
play a crucial role in BBO optimizers~\cite{DE-DDQN,GSF-DQN,AOSsurvey}.
Inspired by the success of recent work~\cite{Lu2020A,chen2024symbol}, we focus on lifelong learning for constructing solution updating rules. 
Specifically, we take each sequentially arrived problem distribution as a distinct learning task and train a MetaBBO model sequentially on those tasks, with the aim of automatically constructing solution updating rules by the model to effectively solve problems from any learned distribution (i.e., tasks).

To achieve this, we propose LiBOG, which consists of three levels of design. Firstly, LiBOG models the lifelong learning process of MetaBBO as a non-stationary MDP~\cite{surveyCRL}. 
% in which each problem distribution is a task, represented by a stationary MDP. 
The non-stationary MDP is formed by a sequence of stationary MDPs, each representing a task to learn (c.f., Section~\ref{sec:MDP}). 
It facilitates the utilization of advanced RL methods. Based on the formulation, the second level (c.f., Section~\ref{sec:SEL_PPO}) focuses on learning to construct solution updating rules within a single task, without addressing catastrophic forgetting. We construct this level following SYMBOL~\cite{chen2024symbol}, the SOTA MetaBBO method for solution updating rule construction. Finally, at the highest level, LiBOG introduces two consolidation mechanisms to address inter-task (c.f., Section~\ref{sec:inter}) and intra-task (c.f., Section~\ref{sec:intra}) catastrophic forgetting, respectively. Figure~\ref{fig:LiBOG} demonstrates the overall lifelong learning process of LiBOG. A detailed pseudocode of LiBOG's learning process
% is presented in Appendix~\ref{ap:LiBOG}.
can be found in the supplementary material.

\subsection{Formulation of Lifelong Learning MetaBBO\label{sec:MDP}}

% Firstly, we 
% take each problem distribution as a task, and 
% formulate the optimization process of a problem 
% sampled from a given distribution 

A BBO problem distribution $P$, i.e., a task, is represented as an MDP, represented by the tuple $\mathcal{M}= \langle \mathcal{S},\mathcal{A},\mathcal{R},\mathcal{P},\mathcal{H},\rho \rangle$. $\mathcal{S},\mathcal{A},\mathcal{R},\mathcal{P}$ are the optimization state space, solution updating rule space, reward function, and state transition function, respectively. $\mathcal{H}$ represents the maximal number of optimization iterations. $\rho$ is the distribution of the initial state, defined by the initial solutions. To solve a sampled problem of the task, an optimizer first generates an initial solution set, forming the initial state following $\rho$. 
Then the optimizer iteratively observes the current state $s\in \mathcal{S}$, and constructs an updating rule $a$ from $\mathcal{A}$ to update the incumbent solutions, which transits $s$ to the next state $s'\in \mathcal{S}$ following $\mathcal{P}$, and provides a reward value $r$ based on $\mathcal{R}$.
% for a MetaBBO method to update the optimizer. 
The iteration will continue for $\mathcal{H}$ times, and then the best-so-far solution will be output.

Specifically, LiBOG represents a state $s$ by a vector of fitness landscape analysis (FLA) metrics~\cite{SURVEY_FLA},  including the distances of decision variable values and objective values of incumbent solutions. Each FLA metrics are normalized so that all tasks share the same state space.
% Appendix~\ref{ap:state} presents details about state representation.
Details about state representation can be found in the supplementary material.
RL-based MetaBBO methods have demonstrated that those FLA metrics are effective in representing the characteristics of optimization states~\cite{DE-DDQN,Lu2020A,GSF-DQN}.
An action $a$ is a solution updating rule, represented by a tree-structure symbolic equation. Section~\ref{sec:SEL_PPO} details the construction process of a solution updating rule, i.e., an action.

% Following~\cite{chen2024symbol}, the reward function is defined as the sum of two terms, $R_f$ and $R_e$, i.e., $\mathcal{R}(\cdot)=\mathcal{R}_f(\cdot)+\mathcal{R}_e(\cdot)$. $\mathcal{R}_f$ is designed to encourage the improvement of objective value (assuming minimization problem), calculated as 
% \begin{equation} \label{eq:exploration_r}
% \mathcal{R}_f(\tau, k) = \frac{f(x^{*,(k)})-f^{opt}}{f(x^{*,(0)})-f^{opt}},
% \end{equation}
% where $\tau=\{(s_t,a_t)\}_{t=0}^{\mathcal{H}}$ is an experience trajectory of the  optimization process, $f(x^{*,(k)})$ is the objective value of the best found solution $x^{*,(k)}$ within optimization iterations $[0,k]$ in $\tau$, and $f^{opt}$ is the known best or optimal objective value of the problem.
% $\mathcal{R}_g$ is designed to guide the learning by a given human-designed optimizer, by encouraging generating
% solutions $X$ with variable values similar to those produced by a given manually designed optimizer $X'$ when applied to the same incumbent solutions.
% \begin{equation} \label{eq:guide_r}
% \mathcal{R}_g(\tau,k)= \frac{\max_{x \in X} \left( \min_{x'\in X'}(||x-x'||_2) \right)}{x_{ub}-x_{lb}},
% \end{equation}
% where $x_{ub}$ and $x_{lb}$ are the upper and lower bound of variable values, i.e., the range of search space. 

The reward function in~\cite{chen2024symbol} is used:
% \begin{equation} \label{eq:exploration_r}
% \mathcal{R}(\tau, k) = \frac{f(x^{*,(k)})-f^{opt}}{f(x^{*,(0)})-f^{opt}} +  \frac{\max_{x \in X} \left( \min_{x'\in X'}(||x-x'||_2) \right)}{x_{ub}-x_{lb}},
% \end{equation}
% \begin{equation} \label{eq:r}
% \mathcal{R}(k) = \frac{f(x^{*,(k)})-f^{opt}}{f(x^{*,(0)})-f^{opt}} \\ +  \frac{\max_{x \in X} \left( \min_{x'\in X'}(||x-x'||_2) \right)}{x_{ub}-x_{lb}},
% \end{equation}
% \begin{align}\label{eq:r}
% \mathcal{R}(k) = & \frac{\max_{x \in X} \left( \min_{x'\in X'}(||x-x'||_2) \right)}{x_{ub}-x_{lb}} \nonumber \\ &+ \frac{f(x^{*,(k)})-f^{opt}}{f(x^{*,(0)})-f^{opt}} ,
% \end{align}
\begin{equation} \label{eq:r}
\mathcal{R}(k) = \frac{f(x^{*,(k)})-f^{opt}}{f(x^{*,(0)})-f^{opt}} \\ +  \frac{d(X_k,X'_k)}{x_{ub}-x_{lb}},
\end{equation}
% where $\tau=\{(s_t,a_t)\}_{t=0}^{\mathcal{H}}$ is an experience trajectory of the  optimization process, 
where $f(x^{*,(k)})$ is the best-so-far objective value within optimization iterations $[0,k]$, and $f^{opt}$
% is the known optimal objective value of the problem. If the optimum is unknown, it can be approximated by running an existing effective BBO optimizer.
is the optimal objective value, \bc{or the known best objective if the optimum is unknown,} of the problem.
$X_k$ and $X'_k$ are the solution population generated by the optimizer from the MetaBBO and by a given manually designed optimizer, respectively, in $k$th iteration, based on the same $X_{k-1}$. $d(X_k,X'_k)=\max_{x \in X_k} \min_{x'\in X'_k}(\|x-x'\|_2) $ is the distance measure of two solution population. $x_{ub}$ and $x_{lb}$ are the upper and lower bounds of variable values.

% Then We model the lifelong learning process of BBO optimizer generation as a non-stationary MDP~\cite{surveyCRL}, represented by the tuple $\mathbf{M}= \langle \mathcal{S}(t),\mathcal{A}(t),\mathcal{R}(t),\mathcal{P}(t),\mathcal{H}(t),\rho(t) \rangle$. For a specific time point $t$, a stationary MDP, defined by the current problem distribution $P_i$, is available for experience collection. 
% We focus on the scenarios where problem distributions, and subsequently tasks, switch discretely, i.e.,
% % $M(t) \in \{M_1,M_2\dots\}$. 
% $M(t) = M_i \forall t \in [s_i,s_{i+1})$, where $s_i$ and $s_{i+1}$ are switch time points. The task boundaries, i.e., the time point $\{s_i\}$ distribution changing, are assumed available. This aligns with many real-world situations. 
% % For example, in energy scheduling, we can anticipate that the problem distribution will shift due to events like sudden wars or fluctuating energy prices, even if the exact nature of the changes is unknown. This knowledge allows us to explicitly store a trained high-performance model for each previously encountered task.
% For example, in scheduling problems of a manufacturing company, managers can clearly identify the time when seasonal changes, which will lead to changes in distributions of material prices and product demand.

% % Period of all tasks lasting is assumed the same, marked as $N$. Therefore, $M(t) = M_{\lfloor t/N \rfloor}=M_i$ where $M_i$ corresponds the $i$th tasks available when $t\in [t,\lfloor t/N \rfloor) ]$

Based on the above formulation, we further model the lifelong learning process of MetaBBO as a discrete non-stationary MDP with piecewise non-stationary function~\cite{surveyCRL}, represented as $\mathbf{M}=\{\mathcal{M}_0,\mathcal{M}_1,\dots\}$. For a specific time point, only the $\mathcal{M}_i$, corresponding to the current problem distribution $P_i$, is available for the optimizer to interact. 
% We assume the identity of the current task is known,
We assume the identities and boundaries of tasks are known,
which aligns with many real-world situations. 
For example, managers of a manufacturing company can identify the current season in scheduling problems, where seasonal changes in material prices and product demands lead to changes in problem distribution.

% For different tasks, $\mathcal{S},\mathcal{A}$ are designed to be identical. 
Different tasks share the same state space $\mathcal{S}$ and action space $\mathcal{A}$.
We assume all problems share the same optimization iteration budget, therefore, $\mathcal{H}$ is also identical for all tasks. 
% $\mathcal{R},\mathcal{P}$ and $\rho$ will change between tasks, 
Due to the different fitness landscapes of different problem distributions, different tasks have different state transition functions and different distributions of initial states. Besides, the reward values obtained from different tasks given the same state and action will also be different, though the formulation of the reward function is the same across tasks.

% To effectively learn across a sequence of problem distributions without forgetting previously acquired knowledge, LiBOG comprises two levels of design. The first level focuses on generate BBO optimizers by learning from the current task $M_i=M(t)$ at a given time point $t$ . We follow the recent study approach of automatically designing BBO optimizers~\cite{chen2024symbol} in this level and employ symbolic equation learning and proximal policy optimization (PPO)~\cite{PPO} to train the optimizer generator (c.f. Section \ref{sec:SEL_PPO}), given their success in automatically designing BBO optimizers that outperform those crafted by human experts. Based on the first level, the second level addresses catastrophic forgetting of knowledge learned from previous tasks $\{M_i,\dots,M_{i-1}\}$, and of knowledge learned at previous time $t'$ from the current task $\{t'|t'<t, M(t')=M_i\}$. An inter-task consolidation mechanism (c.f. Section \ref{sec:inter}) and an intra-task consolidation mechanism (c.f. Section \ref{sec:intra}) are involved in the second level. Fig~\ref{fig:LiBOG} demonstrates the overall framework of LiBOG.
% A more detailed pseudocode of LiBOG is presented in Appendix~\ref{ap:LiBOG} 

\subsection{Symbolic Updating Rule Construction\label{sec:SEL_PPO}}

% \cite{AOSsurvey}
Solution updating rules can be formulated as equations that calculate the variable values of the new solution(s) based on the incumbent solutions (as real-value vectors). For example, the \textit{DE/best/1} rule of DE, a classic human-designed BBO optimizer, can be formulated as $x'=x_{best}+F(x_{1}-x_{2})$, where $x'$ is the new solution, $x_{best}$ is the best incumbent solution, $F$ is a pre-defined parameter, $x_{1}$ and $x_{2}$ are two randomly selected incumbent solutions. Symbolic equation learning facilitates the construction of rules as tree-structure symbolic equations~\cite{zheng2022symbolic,chen2024symbol,NEURIPS2023chen}, where terminal nodes are operands like $x_{best}$ and internal nodes represent operators like $+$.

Following~\cite{chen2024symbol}, we apply a \bc{long short-term memory (LSTM)} network as the policy model to generate solution updating rules, i.e., actions.
% , and PPO algorithm to train the model given a stationary MDP $M(t)$ at time $t$. 
% LSTM is used as the policy model to provide actions.
For each state, the LSTM predicts multiple times sequentially to construct a tree. For each prediction, the model takes FLA metrics, i,e., the state, and the vectorized tree embedding of the current tree as input, and outputs the next node to be added into the tree. Detailed settings of the tree construction can be found in the supplementary material.
% The detailed process of LSTM to construct symbolic equations can be found in Appendix~\ref{ap:LSTM}. 
% forming a rule space that can cover human-designed rules of many classic optimizers. LSTM is applied as the policy model to add nodes iteratively to construct updating rules. The details can be found in Appendix~\ref{}.
Proximal policy optimization (PPO)~\cite{PPO} is used in training, given its success in training to generate BBO optimizers that outperform those crafted by human experts~\cite{chen2024symbol}.

\subsection{Inter-Task Consolidation\label{sec:inter}}

% To preserve knowledge learned from previous tasks, LiBOG prevents significant deviation from the parameters optimized for earlier tasks.
Preventing significant deviation from the parameters optimized for earlier tasks has the potential to preserve learned knowledge.
It can be achieved by applying L2 regularization on the model's parameters. However, equally regularizing all parameters leads to the loss of plasticity~\cite{ewc}.
% For inter-task consolidation, LiBOG adopts the classic elastic weight consolidation (EWC)~\cite{} method. 
In contrast, LiBOG 
adaptively adjusts the regularization strength for each parameter based on its importance to previous tasks, based on elastic weight consolidation (EWC) method~\cite{ewc}. Parameters deemed more critical for earlier tasks are preserved to a greater extent, while less important parameters are allowed larger updates, providing more flexibility for learning new tasks. 
% LiBOG employs elastic weight consolidation (EWC)~\cite{ewc} for inter-task consolidation, quantifying 
The importance of parameters for the $i$th task 
is calculated as:
% as Equation~(\ref{eq:fisher}).
\begin{equation}\label{eq:fisher}
   \Omega^\theta_i = \frac{1}{|\mathcal{J}|} \sum_{\tau \in \mathcal{J}} \left( \frac{1}{|\tau|} \sum_{k=0}^{|\tau|-1} \nabla_\theta \ell_{\theta,k} \nabla_\theta \ell_{\theta,k}^\top \right),
\end{equation}
where $\mathcal{J} = \{\tau_0,\tau_1,\dots\}$ is a given set of experience trajectories, $\tau=\{(s_k, a_k,r_k)\}_{k=0}^{\mathcal{H}}$ is a trajectory obtained by model $\pi_\theta$ from $i$th task, i.e., records of all decisions in an optimization process, and $\ell_{\theta,k} = \log \pi_\theta(a_k | s_k)$ corresponds the action probability generated by model $\pi_\theta$ given $(s_k,a_k) \in \tau$.

Specifically, after completing training on each task, LiBOG records the parameter values $\theta^*$ and calculates the importance $\Omega$ to this task with the trajectories obtained for the last training epoch. During training on a new task, an inter-task consolidation term $\mathcal{CL}_{inter}$ is incorporated into the loss function, calculated as:
\begin{equation}\label{eq:ewc}
   \mathcal{CL}_{inter}(\theta) = \frac{1}{i} \sum_{j=0}^{i-1} (\theta - \theta^*_j)^\top \Omega_j (\theta - \theta^*_j).
\end{equation}
A smaller $\mathcal{CL}_{inter}$ indicates the parameters important to previous tasks are similar to the best value obtained on those tasks.
Consolidation in parameter level is relatively storage-efficient, as it only requires saving the parameters and importance with the space complexity as $O(|\theta|\cdot I)$ rather than retaining any task-specific experiences, where $|\theta|$ is the number of model parameters and $I$ is the total number of tasks.

% \begin{equation}\label{eq:ewc}
%\mathcal{CL}_{inter}(\theta) = \frac{1}{i-1} \sum_{j\in \{0,\dots,i-1\}} (\theta - \theta^*_j)^\top \Omega_j (\theta - \theta^*_j).
% \end{equation}

\subsection{Intra-Task Consolidation\label{sec:intra}}

% Catastrophic forgetting arises from changes in data distribution. Most research on catastrophic forgetting focuses on distributional shifts caused by task transitions. However, in RL, changes in data distribution can also occur within a single task.
% Typically, training data in RL is the experience generated through iterative interactions between the policy model and a given environment,
% leading to temporal correlations among consecutive samples. 
Typically, the distribution of obtained experience in RL could shift due to changes in policy caused by the model parameter updating. This results in intra-task catastrophic forgetting, which can hinder model convergence and reduce generalizability~\cite{igl2021transient}. Common approaches in RL, such as extensive resampling and large experience memory, address this issue but significantly increase storage requirements and computational costs~\cite{lan2023memoryefficient}. Given the inherently computationally expensive nature of BBO, such brute-force methods become less practical.

Existing studies about intra-task forgetting focus on single-task scenarios~\cite{Ghiassian2020improving,pan2021fuzzy,zhang2023catastrophic}, with limited attention to its effects in multi-task settings. We focus on lifelong learning with multiple tasks. 
% In lifelong learning MetaBBO scenarios with multiple tasks, 
Inter-task and intra-task forgetting could interact, leading to more pronounced effects on the training process. Additionally, the substantial stochasticity inherent in the BBO process introduces significant fluctuations in obtained experiences, potentially exacerbating the impacts.

% Distilling the learned knowledge of previous models to the currently updating model by knowledge distillation holds promise for addressing intra-task forgetting~\cite{igl2021transient,zhang2023catastrophic}. However, periodic distillation can be computationally expensive and may restrict learning when the model achieves significant performance improvements between distillation cycles. 

To address the challenges, we propose the elite behavior consolidation (EBC) method. EBC maintains an elite model, parameterized by $\theta_e$, as a reference and updates it as a copy of the new model whenever the new model has better performance than the elite model. In the context of MetaBBO, the model that obtain a better average final objective value on the current tasks is defined as better. During each model update, EBC regularizes the current model's behavior to align with the elite model. Specifically, EBC introduces the KL divergence between the action probability distributions of the current model and the elite model for given states as a loss term $\mathcal{CL}_{intra}$, calculated as: 
\begin{equation}\label{eq:ebc}
\mathcal{CL}_{intra}(\theta) = \frac{1}{|\mathcal{J}|} \sum_{\tau\in \mathcal{J}}\sum_{s \in \tau} \sum_{a\in \mathcal{A}} \pi_{\theta}(a|s) \log \frac{\pi_{\theta}(a|s)}{\pi_{\theta_e}(a|s)}.
\end{equation}
% \begin{equation}\label{eq:ebc}
%    \mathcal{CL}_{intra}(\theta) = \frac{1}{|\mathcal{J}|} \sum_{\tau\in |\mathcal{J}|}\sum_{k=0}^{|\tau|-1} KL(\pi_{\theta}(s_k)||\pi_{\theta_e}(s_k)),
% \end{equation} 
% $KL(\theta\|\theta_e)$ is used instead of $KL(\theta_e\|\theta)$ as it could better prevent $\theta$ from selecting unpromising actions that $\theta_e$ would take with very low (even zero) probability and stabilizing the learning, which is considered important in BBO optimizer generation. 
% % This has also shown good effectiveness in some recent studies~\cite{Balance,2024RKLDarxiv}.

When training on one task is finished, the record of $\theta_e$ will be discarded.
Compared to regularizing model parameters, EBC directly regularizes model behavior, effectively addressing catastrophic forgetting caused by changes in behavioral patterns. It may also better accommodate exploration within the model parameter space, as different sets of model parameters can correspond to similar, near-optimal policy behaviors.

In summary, to maximize the overall reward $\mathcal{R}$ while addressing both inter-task and intra-task forgetting, LiBOG updates a model on each task with the following loss function.
\begin{equation} \label{eq:over_loss}
\mathcal{L}(\cdot) =  \mathcal{L}_{PPO}(\cdot) + \alpha \cdot \mathcal{CL}_{inter}(\cdot) + \beta \cdot \mathcal{CL}_{intra}(\cdot),
\end{equation}
where $\mathcal{L}_{PPO}$ is the loss function of PPO algorithm to maximize $\mathcal{R}$, $\alpha$ and $\beta$ are two pre-defined hyper-parameters balancing the two consolidation terms.

\section{Experiments}

We focus on the MetaBBO scenario where various problem distributions sequentially arise.
Through experimental studies\footnote{Our code is available in
https://github.com/PeiJY/LiBOG.}, we aim to answer the following research questions.

\begin{itemize}
\item \textbf{Optimization effectiveness:} Does LiBOG demonstrate superior optimization ability on learned problems compared to SOTA human-designed optimizers and MetaBBO methods without lifelong learning?
\item \textbf{Addressing catastrophic forgetting:} What are the effects of catastrophic forgetting in such scenarios, and are the two consolidation mechanisms in LiBOG effective in mitigating this issue?
\item \textbf{Hyper-parameter sensitivity:} How sensitive is LiBOG to the weights of consolidation terms, i.e., $\alpha$ and $\beta$?
\end{itemize}

\paragraph{Problem Distributions.}

The training dataset is constructed from the widely studied IEEE CEC Numerical Optimization Competition Benchmark~\cite{cec2021so}. 
% It features a diverse set of challenging synthetic problems designed to mimic real-world optimization complexities. These problems encompass various global properties, such as uni-modal and multi-modal landscapes, (non-)separability, and (a)symmetry, along with intricate local features like rugged landscapes, flat regions, and discontinuities. 
The benchmark is known for its extensive use in the optimization research community, serving for comparative studies and fostering advancements in BBO methods.
Four function categories, namely, uni-modal, basic, hybrid, and composition, are provided. 
Different categories own different properties and landscape features, such as uni-modal and multi-modal landscapes, (non-)separability, and (a)symmetry. 
% A total of ten functions exist in the benchmark.
Each function can be configured by the dimension, searching space range, function offset, and function rotation, forming a specific problem. We set the dimension and searching space to $10$ and $[-100,100]^{10}$ for all the functions. Then, we form each category as a task by introducing a distribution of offset $z\sim U[-80,80]^{10}$ and a distribution of rotation uniformly distributed in  $\mathbb{R}^{10\times 10}$ for each function of the category. When sampling a problem from a task, all functions of the corresponding category are selected with the same probability. 
In summary, four tasks $\{P^{U}, P^{B}, P^{H}, P^{C}\}$ are constructed, corresponding to uni-modal, basic, hybrid, and composition categories, respectively.
% , with uniformly random offset distribution and rotation distribution. 
% We randomly generate 3 different orders of tasks for experiments.
% (1) Uni-modal, basic, hybrid, composition. (2) Composition, uni-modal, hybrid, composition. (3) Uni-modal, composition, hybrid, basic.
We randomly generated three different task orders aiming to eliminate the influence of specific task orders on the experimental results. Details about the training dataset and the task orders
% can be found in Appendix~\ref{ap:parameters}.
can be found in the supplementary material.

\paragraph{Baselines.}
To the best of our knowledge, there is no existing work on lifelong learning MetaBBO, as current MetaBBO methods focus on learning for a single task. We compare LiBOG with the SOTA human-designed BBO optimizer MadDE~\cite{MadDE}, which is one of the winners of the CEC competition, and SYMBOL~\cite{chen2024symbol}, a SOTA MetaBBO method for single-task learning. For SYMBOL, we applied two strategies to learn across multiple consecutive tasks: (i) randomly generating a model each time a new task appears and then training the model on the new task, denoted as \textit{restart}, and (ii)
% taking the final model of learning on the previous task as the initial model for the next task, 
directly updating the obtained model on a new task,
denoted as \textit{fine-tuning}.
% and (iii) the classic lifelong learning method EWC (without any intra-task consolidation) denoted as \textit{EWC}. 
Additionally, we compared with a baseline method assuming all functions are available at once and sampling problems from all of them to train SYMBOL in each epoch, denoted as \textit{all-task}.

% \begin{table*}[htbp]
% \centering
% \caption{The rank (smaller better) of test performance on each task of each task order.}
% \begin{tabular}{c|cccc|c|cccc|c|cccc|c}
% \toprule
% \multirow{2}{*}{\textbf{Method}} & \multicolumn{5}{c|}{\textbf{Order 0}}    & \multicolumn{5}{c|}{\textbf{Order 1}}& \multicolumn{5}{c}{\textbf{Order 2}} \\ \cline{2-16}
%    & $P_0$ & $P_1$ & $P_2$ & $P_3$ & avg.& $P_0$ & $P_1$ & $P_2$ & $P_3$ & avg.   & $P_0$ & $P_1$ & $P_2$ & $P_3$ & avg.  \\ 
%    \midrule
% \textbf{LiBOG}   & \textbf{1} & \textbf{1} & \textbf{1} & \textbf{1} & \textbf{1} & \textbf{1} & \textbf{1} & \textbf{1} & 2& \textbf{1.25} & \textbf{1} & 2& 2& \textbf{1} & \textbf{1.5} \\
% \textbf{restart} & 4& 4& 2& 2& 3& 2& 2& 4& \textbf{1} & 2.25& 4& \textbf{1} & \textbf{1} & 3& 2.25    \\
% \textbf{fine-tuning}  & 2& 3& 3& 3& 2.75  & 4& 4& 3& 4& 3.75& 3& 4& 3& 4& 3.5\\
% \textbf{all-task}& 5& 2& 5& 5& 4.25  & 5& 5& 2& 5& 4.25& 5& 5& 5& 2& 4.25    \\
% \textbf{MadDE}& 3& 5& 4& 4& 4& 3& 3& 5& 3& 3.5 & 2& 3& 4& 5& 3.5 \\
% \bottomrule
% \end{tabular}
% \label{tab:overall_result}
% \end{table*}

\begin{table*}[htbp]
\centering
\resizebox{\textwidth}{!}{
\begin{tabular}{c|cccc|c|cccc|c|cccc|c}
\toprule
\multirow{3}{*}{\textbf{Method}} & \multicolumn{5}{c|}{\textbf{Order 0}}& \multicolumn{5}{c|}{\textbf{Order 1}}  & \multicolumn{5}{c}{\textbf{Order 2}} \\ \cline{2-16}
   & {\textbf{$P_0$}} & {\textbf{$P_1$}} & {$P_2$} & {$P_3$} & {\multirow{2}{*}{avg.}} & {$P_0$} & {$P_1$} & {$P_2$} & $P_3$ & {\multirow{2}{*}{avg.}} & {$P_0$} & {$P_1$} & {$P_2$} & {$P_3$} & {\multirow{2}{*}{avg.}} \\
   & ($P^U$)    & \textbf{($P^B$)} & ($P^H)$   & ($P^C$) & {}& ($P^C$) & ($P^U$) & ($P^B$) & ($P^H)$ & {}& ($P^U$) & ($P^C$) & ($P^H)$   & ($P^B$) & {}\\
   \midrule
\textbf{LiBOG} & \textbf{1}   & \textbf{1}   & \textbf{1}& \textbf{1}& \textbf{1}  & \textbf{1}& \textbf{1}& 2   & 2 & \textbf{1.5}& 2   & \textbf{1}& \textbf{1}& 2   & \textbf{1.5}\\
\textbf{restart} & 4    & 4   & 2   & 2   & 3& 4   & 4   & \textbf{1}& \textbf{1} & 2.5& 4   & 5   & 2   & \textbf{1}& 3\\
\textbf{fine-tuning}  & 2    & 3    & 3   & 3   & 2.75   & 3   & 3   & 4   & 4 & 3.5& 3   & 3   & 3   & 4   & 3.25   \\
\textbf{all-task}& 5    & 2    & 5   & 5   & 4.25   & 5   & 5   & 3   & 5 & 4.5& 5   & 4   & 5   & 3   & 4.25   \\
\textbf{MadDE} & 3    & 5    & 4   & 4   & 4& 2  & 2& 5& 3 & 3 & \textbf{1}  & 2& 4& 5& 3 \\
\bottomrule
\end{tabular}
}
\caption{The rank of test results on each task. Smaller ranks indicate better performance.}
\label{tab:overall_result}
\end{table*}

\paragraph{Hyper-parameter Setting and Performance Evaluation.}

For LiBOG, \textit{restart} and \textit{fine-tuning},
% and \textit{EWC},
the models are trained on each task for 100 epochs equally. 
Following~\cite{chen2024symbol}, \textit{all-task} trains the model on all functions simultaneously for 100 epochs. Both SYMBOL and LiBOG use MadDE as the guide optimizer for reward calculation. For LiBOG, the values of $\alpha$ and $\beta$ are set to 1 based on the rule of thumb. 
% For testing the optimization performance of a task, 
To test the optimization performance of an optimizer on a task,
we sample 32 problems with the corresponding distribution. We run the optimizer to solve each of the problems, and take the average output objective value over the problems 
% as the test performance for the function after normalizing based on the optimal objective value and the worst objective value found on the function of all runs. And the average test performance of all functions of a task is used
as the test performance on the task. 
The objective values are normalized between 0 and 1. A larger objective value indicates better optimization performance. 
% $1$ represents the optimal objective value, and $0$ represents the worst objective value found.
More settings about hyper-parameters and the normalization method 
% can be found in Appendix~\ref{ap:parameters} and~\ref{ap:normalization}
can be found in the supplementary material.

\subsection{Effectiveness Evaluation\label{sec:effectiveness}}
To verify the effectiveness of LiBOG in generating high-performance optimizers, we tested and compared the performance of LiBOG and all baseline methods on each task. For LiBOG and \textit{fine-tuning}, the model obtained after training on the final task was used for testing. For \textit{restart}, the model trained on a task is used for testing on that task. For \textit{all-task}, the model obtained after the one-off training was tested. MadDE does not involve a learning process and is directly tested on all functions.
% To evaluate the effectiveness of LiBOG for solving all tasks, we compared the test performance of all baseline methods on each task after completing the learning process over all tasks. 
Each learning method underwent 10 independent learning runs for each task order. The model obtained in each run is tested, and the average test results over the 10 runs are compared. Table~\ref{tab:overall_result} presents the rank of each method. 
More detailed results 
% can be found in Appendix~\ref{ap:results}.
can be found in the supplementary material.
% The box figure in Figure~\ref{fig:overall_result} summarizes the test performance on each task. Each box contains 10 results, corresponding to the 10 runs. 

% At the function level, LiBOG achieved the best test performance on seven out of ten functions. At the task level, LiBOG outperformed all other methods, including the SOTA human-designed optimizer (i.e., MadDE), across every task.
LiBOG outperforms all baseline methods for each of the three task orders.
Specifically, LiBOG's superior performance compared to \textit{fine-tuning} demonstrates the effectiveness of LiBOG's two consolidation methods in addressing catastrophic forgetting and maintaining good plasticity. \textit{Restart} achieved the best performance in some cases, as expected due to its task-specific training and lack of influence from task distribution differences. 
% However, LiBOG outperformed \textit{restart} in average rank, 
However, LiBOG is ranked in first place for eight of the twelve cases over the three orders,
indicating that, in general, LiBOG effectively transfers knowledge of previously learned tasks to enhance the learning of subsequent tasks without forgetting, though the problem distributions are different.

The rank stability of LiBOG across different task orders (average ranks of 1, 1.5, and 1.5) demonstrates its robustness to changes in task orders.

\subsection{Addressing Forgetting\label{sec:forgetting}}

\begin{figure}[htbp]
\centering
\includegraphics[width=.9\linewidth]{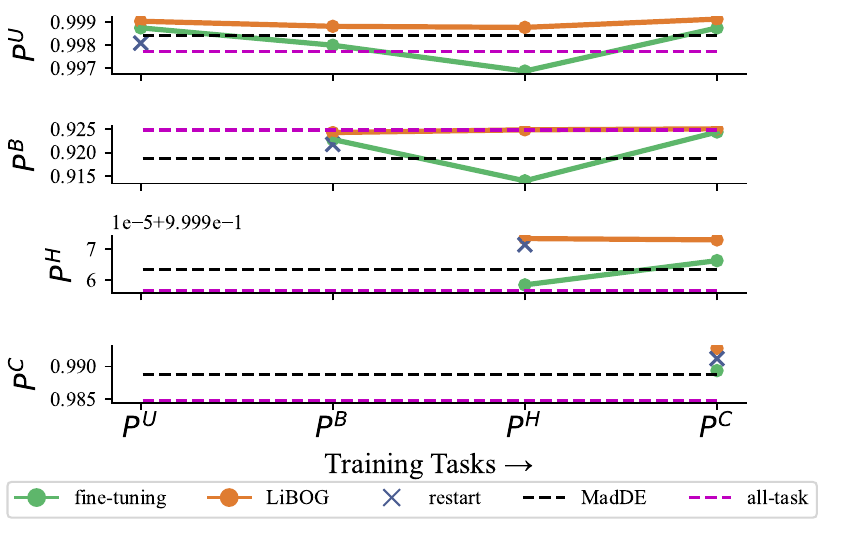}
\caption{Test performance on each learned task during the lifelong learning process of task order 0. In \textit{fine-tuning}, catastrophic forgetting of previous tasks is significant, but is mild in LiBOG.}
\label{fig:forgetting}
\end{figure}

We analyzed the impact of catastrophic forgetting in the studied lifelong learning MetaBBO scenarios, and the effectiveness of LiBOG in addressing catastrophic forgetting. Figure~\ref{fig:forgetting} shows the test performance on previously learned tasks after the training of each new task of one task order. Though there is no lifelong learning process in MadDE, \textit{all-task} and \textit{restart}, we add them for reference. Results of other orders demonstrate a similar pattern, and 
% can be found in Appendix~\ref{ap:results}.
can be found in the supplementary material.

% \textit{Fine-tuning} exhibits significant catastrophic forgetting, while, with the consolidations, LiBOG shows much less forgetting of previously learned tasks. Even when slight forgetting occurs, LiBOG consistently outperforms the human-expert-designed MadDE. 
\textit{Fine-tuning} shows a significant performance drop on \bc{the first task $P_0$ (i.e., uni-modal functions $P^U$) and the second task $P_1$ (i.e., basic functions $P^B$)} after training on the subsequent two and one tasks, respectively, performing worse than \textit{all-task} and MadDE, indicating a significant catastrophic forgetting. In contrast, LiBOG experiences very small performance degradation after learning new tasks, suggesting that forgetting in LiBOG is mild.
In some cases, training the model on a task (e.g., $P_3$ ($P_C$) in Figure~\ref{fig:forgetting}) was observed to enhance its performance on previously learned tasks.
This is perhaps due to the inherent function similarity between tasks. 
This explains why LiBOG outperforms others, as it effectively utilizes similar tasks to better learn shared knowledge while retaining knowledge from dissimilar tasks without forgetting.

\begin{figure}[htbp]
\centering
\includegraphics[width=.9\linewidth]{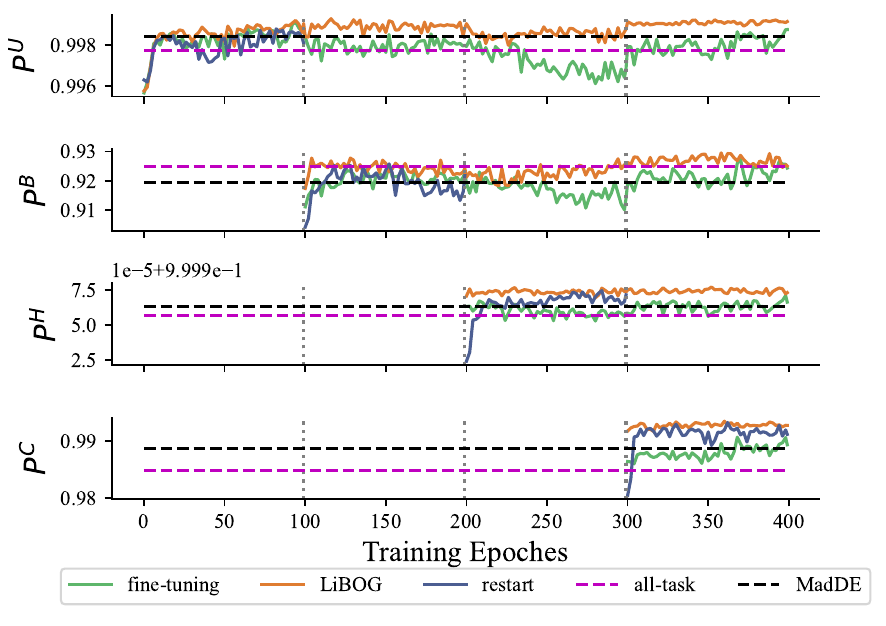}
\caption{Test performance on each task of models obtained during the lifelong learning process, under task order 0. Vertical gray lines indicate the time of task changes. }
\label{fig:training_curve}
\end{figure}

To further analyze the training process within each task, we recorded the models obtained after each training epoch throughout the lifelong learning process and tested them on the current and previous tasks. Figure~\ref{fig:training_curve} shows the results in task order 0. 
% Appendix~\ref{ap:results} provides the results of the other two orders.
The results of the other two orders can be found in the supplementary material.

% We observe that intra-task forgetting significantly affects multi-task scenarios as well. When the policy model trained on a previous task is used as the initial model for the next task, differences between tasks can lead to substantial changes in experience distribution between the early and late stages of training. This exacerbates intra-task forgetting, reducing performance on the current task. Additionally, significant shifts in experience distribution, combined with the inherent stochasticity of the BBO process, cause drastic updates to model parameters, thereby intensifying inter-task forgetting and degrading performance on previously learned tasks.

% Restart begins learning on each task from a randomly generated model, showing expected performance improvements on each task. 
The performance of both \textit{fine-tuning} and LiBOG on $P_1$ ($P^B$), $P_2$ ($P^H$), and $P_3$ ($P^C$) is significantly better than \textit{restart} on epochs 100, 200, and 300, respectively.
% \textit{Fine-tuning} and LiBOG, with the model learned from the previous as the initial model for a new task, start with better initial performance compared to \textit{restart}. 
It indicates that using the model trained on previous tasks as the initial model leads to better initial performance, compared to a randomly initialized model used in \textit{restart}. Helpful knowledge is transferred with the trained model.

% However, in \textit{fine-tuning}, as the number of tasks increases, the accumulated disruptions caused by distribution changes degrade the initial performance of \textit{fine-tuning}. By the final task, the initial performance of \textit{fine-tuning} becomes almost as poor as that of a random model.
In some cases, during training on a new task (e.g, training on $P_2$ ($P^H$) during epochs 200-299 in Figure~\ref{fig:training_curve}), \textit{fine-tuning} gradually loses performance on previous tasks (e.g., performance reducing on $P_0$ ($P^U$) and $P_1$ ($P^B$)), demonstrating significant inter-task forgetting.
Moreover, 
% despite having knowledge from previous tasks and starting with a better initial model, 
\textit{fine-tuning} fails to effectively learn and improve performance on the new task. 
% When the policy model trained on a previous task is used as the initial model for the next task, differences between tasks can lead to substantial changes in experience distribution between the early and late stages of training. This exacerbates intra-task forgetting, reducing performance on the current task. Additionally, 
Significant shifts in experience distribution, combined with the inherent stochasticity of the BBO process, could be the reason, leading to intra-task forgetting and poor plasticity.
% In the context of lifelong learning MetaBBO, even if intra-task forgetting has a limited impact on single-task learning (as shown by the restart results), addressing intra-task forgetting becomes crucial when learning across multiple tasks.

% In LiBOG,
% the inter-task consolidation effectively preserves the knowledge learned from previous tasks, while the intra-task consolidation addresses the challenges posed by changing experience distributions within a single task, stabilizing the learning process on new tasks. 
% These 
% the two consolidation mechanisms work together to facilitate both the retention of old knowledge and the stable learning of new knowledge. 
% As illustrated in Figure~\ref{fig:training_curve}, LiBOG 
% steadily improves the model during training on the first two tasks. 
LiBOG obtained good performance during learning on the earlier tasks (i.e., on $P_0$ ($P^U$) and $P_1$ ($P^B$) during epochs 0-199 in Figure~\ref{fig:training_curve}).
When subsequent tasks are introduced, LiBOG demonstrates promising performance on the new tasks even before dedicated training (i.e., on $P_2$ ($P^H$) and $P_3$ ($P^C$) at epochs 200 and 300, respectively), while maintaining this performance stably without forgetting.
It indicates that the two consolidation mechanisms work together to facilitate both the retention of old knowledge and the stable learning of new knowledge.

\subsection{Sensitivity Analysis and Ablation Study}

We evaluated the performance of LiBOG under different hyper-parameter settings,
% and task-specific problem distributions, 
and conducted an ablation study to provide a more comprehensive and reliable analysis.
% More details about the settings can be found in Appendix~\ref{ap:analysis}.

\begin{figure}
\centering
\includegraphics[width=0.75\linewidth]{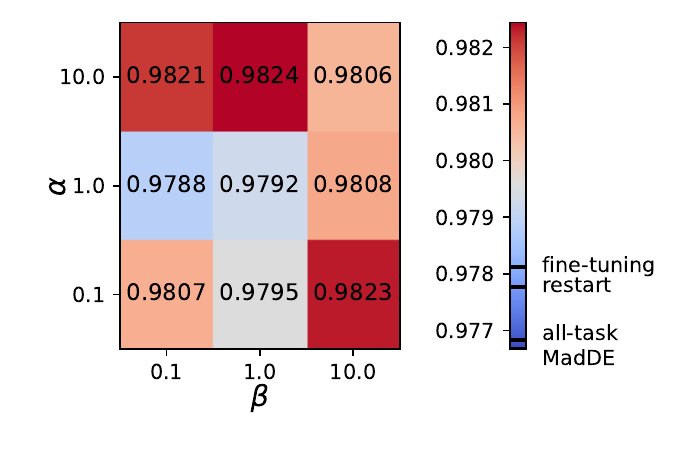}
% \caption{Average performance on all tasks of LiBOG trained with different values of $\alpha$ and $\beta$. Under all settings, LiBOG outperforms baselines.}
\caption{Performance of LiBOG trained with different values of $\alpha$ and $\beta$. Under all settings, LiBOG outperforms baselines.}
\label{fig:weights}
\end{figure}

\paragraph{Consolidation weights.} 
To address catastrophic forgetting, LiBOG incorporates two consolidation mechanisms, each introducing a term to the loss function, controlled by two weights $\alpha$ and $\beta$. Under task order 0, we trained LiBOG with different weight settings and recorded its average test performance across all tasks after learning all tasks. The tested candidate values are $\{0.1,1,10\}$ for both $\alpha$ and $\beta$. Each setting underwent five lifelong learning runs, and Figure~\ref{fig:weights} demonstrates the average of the five runs.  Although the weights affect the performance, the performance of LiBOG is relatively stable, and outperforms baselines for all settings.
% restart and MadDE for eight and all settings, respectively. 
% the weight settings have some impact on LiBOG's performance. However, in most configurations, LiBOG consistently outperforms the human-expert-designed optimizer MadDE after lifelong learning, demonstrating that LiBOG is not highly sensitive to the weight parameter settings.

% \begin{table}[htbp]
% \centering
% \caption{Test performance of LiBOG after learning with different settings of consolidation losses' weights, $\alpha$ and $\beta$.}
% \begin{tabular}{c|c}
% &  \\
% & 
% \end{tabular}
% \label{tab:weights}
% \end{table}

\paragraph{Ablation study.} 
We conducted an ablation study to evaluate the contribution of the two consolidation mechanisms. By individually removing the inter-task consolidation and intra-task consolidation mechanisms, we created two ablation versions: LiBOG with only intra-task consolidation (denoted as \textit{only-intra}) and LiBOG with only inter-task consolidation (denoted as \textit{only-inter}). For each ablation version, we performed lifelong learning five repeat runs on task order 0 and tested the final model's performance across all tasks. The best-performance weight values above are used. 
Table~\ref{tab:ablation} shows the average of the five runs. LiBOG outperforms both ablation versions, indicating that both consolidation mechanisms are critical to LiBOG's effectiveness, as removing either significantly reduces its performance.

\begin{table}[htbp]
\centering
\begin{tabular}{cc|cc}
\toprule
\textbf{Method}& \textbf{Performance} & \textbf{Method}& \textbf{Performance} \\
\midrule
\textbf{LiBOG} & \textbf{0.982440}    & fine-tuning & 0.978123 \\
only-intra  & 0.982306 & restart& 0.977762\\
only-inter  & 0.975054 & all-task    & 0.976833\\
 &  & MadDE & 0.976669  \\
 \bottomrule
\end{tabular}
\caption{Ablation study results of inter-consolidation and intra-consolidation mechanisms in LiBOG.}
\label{tab:ablation}
\end{table}

\section{Conclusions}
% Existing MetaBBO methods focus on generating efficient BBO optimizers through one-off training on extensive problems sampled from a pre-available stationary distribution. However, 
In real-world scenarios, diverse optimization problems often arise sequentially, with the problem distribution changing over time. 
Focusing on these scenarios, this paper studies the unexplored paradigm of lifelong learning for BBO optimizer generation.
% , where a model is trained across a sequence of tasks corresponding to different problem distributions. 
% The goal is to generate BBO optimizers with high performance on all learned tasks while transferring knowledge to improve future problem-solving.
% We model the lifelong learning process of MetaBBO as a non-stationary MDP, where the state transition, reward function, and initial state distribution vary across tasks. Then we propose LiBOG, a novel lifelong RL-based approach for BBO optimizer generation. 
We propose LiBOG, a novel lifelong learning-based MetaBBO approach. In LiBOG, the optimization process of MetaBBO is formulated as a non-stationary MDP, where the state transition, reward distribution, and initial state distribution vary across tasks.
LiBOG employs elastic weight consolidation (EWC) for inter-task consolidation, mitigating catastrophic forgetting caused by differences in problem distributions between tasks. Additionally, we propose elite behavior consolidation (EBC), a novel method that aligns model behavior with elite models obtained within a single task 
% to address intra-task forgetting.
for intra-task consolidation.
% LiBOG uses EBC for intra-task consolidation.
% Extensive e
Experiments on various task orders and hyper-parameter settings demonstrate the effectiveness and robustness of LiBOG in transferring knowledge for enhanced learning on new tasks and addressing catastrophic forgetting. The ablation study verifies the contribution of each consolidation mechanism.

Despite its promising performance, LiBOG has some potential limitations. It cannot be directly applied to scenarios with continuously changing problem distribution or the distribution changing time points are unknown (i.e., unknown task boundaries), as EWC requires explicitly storing a model and parameter importance matrix for each task. Furthermore, EBC relies on constraining model updates based on elite models, which may limit performance when the performance landscape is highly complex and requires extensive exploration. Addressing these limitations will have the potential to enhance LiBOG's expertise in broader applications.

%% The file named.bst is a bibliography style file for BibTeX 0.99c
\clearpage
\bibliographystyle{named}

\bibliography{ijcai25}

\begin{thebibliography}{}

\bibitem[\protect\citeauthoryear{Abel \bgroup \em et al.\egroup }{2018}]{Abel2018policy}
David Abel, Yuu Jinnai, Sophie~Yue Guo, George Konidaris, and Michael Littman.
\newblock Policy and value transfer in lifelong reinforcement learning.
\newblock In {\em International Conference on Machine Learning}, volume~80 of {\em Proceedings of Machine Learning Research}, pages 20--29. PMLR, 10--15 Jul 2018.

\bibitem[\protect\citeauthoryear{Audet and Kokkolaras}{2016}]{Audet_2016}
Charles Audet and Michael Kokkolaras.
\newblock Blackbox and derivative-free optimization: theory, algorithms and applications.
\newblock {\em Optimization and Engineering}, 17(1):1–2, February 2016.

\bibitem[\protect\citeauthoryear{Beyer and Schwefel}{2002}]{Beyer_2002}
Hans-Georg Beyer and Hans-Paul Schwefel.
\newblock Evolution strategies -- {A} comprehensive introduction.
\newblock {\em Natural Computing}, 1(1):3--52, 2002.

\bibitem[\protect\citeauthoryear{Bi \bgroup \em et al.\egroup }{2022}]{bbometal}
Sirui Bi, Benjamin Stump, Jiaxin Zhang, Yousub Lee, John Coleman, Matt Bement, and Guannan Zhang.
\newblock Blackbox optimization for approximating high-fidelity heat transfer calculations in metal additive manufacturing.
\newblock {\em Results in Materials}, 13:100258, 2022.

\bibitem[\protect\citeauthoryear{Biswas \bgroup \em et al.\egroup }{2021}]{MadDE}
Subhodip Biswas, Debanjan Saha, Shuvodeep De, Adam~D Cobb, Swagatam Das, and Brian~A Jalaian.
\newblock Improving differential evolution through {B}ayesian hyperparameter optimization.
\newblock In {\em IEEE Congress on Evolutionary Computation}, pages 832--840, 2021.

\bibitem[\protect\citeauthoryear{Chaybouti \bgroup \em et al.\egroup }{2022}]{chaybouti2022metalearning}
Sofian Chaybouti, Ludovic~Dos Santos, Cedric Malherbe, and Aladin Virmaux.
\newblock Meta-learning of black-box solvers using deep reinforcement learning.
\newblock In {\em Sixth Workshop on Meta-Learning at the Conference on Neural Information Processing Systems}, 2022.

\bibitem[\protect\citeauthoryear{Chen \bgroup \em et al.\egroup }{2017}]{chen2017ICML}
Yutian Chen, Matthew~W. Hoffman, Sergio~G{\'o}mez Colmenarejo, Misha Denil, Timothy~P. Lillicrap, Matt Botvinick, and Nando de~Freitas.
\newblock Learning to learn without gradient descent by gradient descent.
\newblock In {\em International Conference on Machine Learning}, volume~70 of {\em Proceedings of Machine Learning Research}, pages 748--756. PMLR, 06--11 Aug 2017.

\bibitem[\protect\citeauthoryear{Chen \bgroup \em et al.\egroup }{2023}]{NEURIPS2023chen}
Xiangning Chen, Chen Liang, Da~Huang, Esteban Real, Kaiyuan Wang, Hieu Pham, Xuanyi Dong, Thang Luong, Cho-Jui Hsieh, Yifeng Lu, and Quoc~V Le.
\newblock Symbolic discovery of optimization algorithms.
\newblock In {\em Advances in Neural Information Processing Systems}, volume~36, pages 49205--49233. Curran Associates, Inc., 2023.

\bibitem[\protect\citeauthoryear{Chen \bgroup \em et al.\egroup }{2024}]{chen2024symbol}
Jiacheng Chen, Zeyuan Ma, Hongshu Guo, Yining Ma, Jie Zhang, and Yue-Jiao Gong.
\newblock {SYMBOL}: Generating flexible black-box optimizers through symbolic equation learning.
\newblock In {\em International Conference on Learning Representations}, 2024.

\bibitem[\protect\citeauthoryear{Ghiassian \bgroup \em et al.\egroup }{2020}]{Ghiassian2020improving}
Sina Ghiassian, Banafsheh Rafiee, Yat~Long Lo, and Adam White.
\newblock Improving performance in reinforcement learning by breaking generalization in neural networks.
\newblock In {\em International Conference on Autonomous Agents and MultiAgent Systems}, page 438–446, Richland, SC, 2020. International Foundation for Autonomous Agents and Multiagent Systems.

\bibitem[\protect\citeauthoryear{Gomes \bgroup \em et al.\egroup }{2021}]{gomes2021meta}
Hugo~Siqueira Gomes, Benjamin Léger, and Christian Gagné.
\newblock Meta learning black-box population-based optimizers, 2021.

\bibitem[\protect\citeauthoryear{Gu \bgroup \em et al.\egroup }{2021}]{gu2021optimizing}
Bin Gu, Guodong Liu, Yanfu Zhang, Xiang Geng, and Heng Huang.
\newblock Optimizing large-scale hyperparameters via automated learning algorithm.
\newblock {\em arXiv, 2102.09026}, 2021.

\bibitem[\protect\citeauthoryear{Handoko \bgroup \em et al.\egroup }{2014}]{handoko2014RL}
Stephanus~Daniel Handoko, Duc~Thien Nguyen, Zhi Yuan, and Hoong~Chuin Lau.
\newblock Reinforcement learning for adaptive operator selection in memetic search applied to quadratic assignment problem.
\newblock In {\em Companion Publication of Conference on Genetic and Evolutionary Computation}, page 193–194. ACM, 2014.

\bibitem[\protect\citeauthoryear{Hoos}{2012}]{hoos2012automated}
Holger~H Hoos.
\newblock Automated algorithm configuration and parameter tuning.
\newblock In {\em Autonomous search}, pages 37--71. Springer, 2012.

\bibitem[\protect\citeauthoryear{Hussain \bgroup \em et al.\egroup }{2018}]{kashif2018metaheuristic}
Kashif Hussain, Mohd Najib~Mohd Salleh, Shi Cheng, and Yuhui Shi.
\newblock Metaheuristic research: A comprehensive survey.
\newblock {\em Artificial Intelligence Review}, 52(4):2191--2233, 2018.

\bibitem[\protect\citeauthoryear{Igl \bgroup \em et al.\egroup }{2021}]{igl2021transient}
Maximilian Igl, Gregory Farquhar, Jelena Luketina, Wendelin Boehmer, and Shimon Whiteson.
\newblock Transient non-stationarity and generalisation in deep reinforcement learning.
\newblock In {\em International Conference on Learning Representations}, 2021.

\bibitem[\protect\citeauthoryear{Khetarpal \bgroup \em et al.\egroup }{2022}]{surveyCRL}
Khimya Khetarpal, Matthew Riemer, Irina Rish, and Doina Precup.
\newblock Towards continual reinforcement learning: A review and perspectives.
\newblock {\em Journal of Artificial Intelligence Research}, 75:1401–1476, December 2022.

\bibitem[\protect\citeauthoryear{Kirkpatrick \bgroup \em et al.\egroup }{1983}]{10.1126/science.220.4598.671}
S.~Kirkpatrick, C.~D. Gelatt, and M.~P. Vecchi.
\newblock Optimization by simulated annealing.
\newblock {\em Science}, 220(4598):671--680, 1983.

\bibitem[\protect\citeauthoryear{Kirkpatrick \bgroup \em et al.\egroup }{2017}]{ewc}
James Kirkpatrick, Razvan Pascanu, Neil Rabinowitz, Joel Veness, Guillaume Desjardins, Andrei~A. Rusu, Kieran Milan, John Quan, Tiago Ramalho, Agnieszka Grabska-Barwinska, Demis Hassabis, Claudia Clopath, Dharshan Kumaran, and Raia Hadsell.
\newblock Overcoming catastrophic forgetting in neural networks.
\newblock {\em Proceedings of the National Academy of Sciences}, 114(13):3521--3526, 2017.

\bibitem[\protect\citeauthoryear{Lan \bgroup \em et al.\egroup }{2023}]{lan2023memoryefficient}
Qingfeng Lan, Yangchen Pan, Jun Luo, and A.~Rupam Mahmood.
\newblock Memory-efficient reinforcement learning with value-based knowledge consolidation.
\newblock {\em Transactions on Machine Learning Research}, 2023.

\bibitem[\protect\citeauthoryear{Lechner \bgroup \em et al.\egroup }{2022}]{pyhopper}
Mathias Lechner, Ramin Hasani, Philipp Neubauer, Sophie Neubauer, and Daniela Rus.
\newblock Pyhopper -- {H}yperparameter optimization.
\newblock {\em arXiv, 2210.04728}, 2022.

\bibitem[\protect\citeauthoryear{Liu \bgroup \em et al.\egroup }{2023}]{howgoodNCO}
Shengcai Liu, Yu~Zhang, Ke~Tang, and Xin Yao.
\newblock How good is neural combinatorial optimization? {A} systematic evaluation on the traveling salesman problem.
\newblock {\em IEEE Computational Intelligence Magazine}, 18(3):14--28, 2023.

\bibitem[\protect\citeauthoryear{Lu \bgroup \em et al.\egroup }{2020}]{Lu2020A}
Hao Lu, Xingwen Zhang, and Shuang Yang.
\newblock A learning-based iterative method for solving vehicle routing problems.
\newblock In {\em International Conference on Learning Representations}, 2020.

\bibitem[\protect\citeauthoryear{Ma \bgroup \em et al.\egroup }{2023}]{NEURIPS2023_232eee8e}
Zeyuan Ma, Hongshu Guo, Jiacheng Chen, Zhenrui Li, Guojun Peng, Yue-Jiao Gong, Yining Ma, and Zhiguang Cao.
\newblock Metabox: A benchmark platform for meta-black-box optimization with reinforcement learning.
\newblock In {\em Advances in Neural Information Processing Systems}, volume~36, pages 10775--10795. Curran Associates, Inc., 2023.

\bibitem[\protect\citeauthoryear{Malan and Engelbrecht}{2013}]{SURVEY_FLA}
Katherine~M. Malan and Andries~P. Engelbrecht.
\newblock A survey of techniques for characterising fitness landscapes and some possible ways forward.
\newblock {\em Information Sciences}, 241:148--163, 2013.

\bibitem[\protect\citeauthoryear{Mohamed \bgroup \em et al.\egroup }{2021}]{cec2021so}
Ali~Wagdy Mohamed, Anas~A Hadi, Ali~Khater Mohamed, Prachi Agrawal, Abhishek Kumar, and PN~Suganthan.
\newblock Problem definitions and evaluation criteria for the {CEC} 2021 on single objective bound constrained numerical optimization.
\newblock In {\em IEEE Congress on Evolutionary Computation}, 2021.

\bibitem[\protect\citeauthoryear{Pan \bgroup \em et al.\egroup }{2021}]{pan2021fuzzy}
Yangchen Pan, Kirby Banman, and Martha White.
\newblock Fuzzy tiling activations: A simple approach to learning sparse representations online.
\newblock In {\em International Conference on Learning Representations}, 2021.

\bibitem[\protect\citeauthoryear{Pei \bgroup \em et al.\egroup }{2024}]{online-offline-aos}
Jiyuan Pei, Jialin Liu, and Yi~Mei.
\newblock Learning from offline and online experiences: A hybrid adaptive operator selection framework.
\newblock In {\em Proceedings of the Genetic and Evolutionary Computation Conference}, page 1017–1025, New York, NY, USA, 2024. ACM.

\bibitem[\protect\citeauthoryear{Pei \bgroup \em et al.\egroup }{2025}]{AOSsurvey}
Jiyuan Pei, Yi~Mei, Jialin Liu, Mengjie Zhang, and Xin Yao.
\newblock Adaptive operator selection for meta-heuristics: {A} survey.
\newblock {\em IEEE Transactions on Artificial Intelligence}, pages 1--21, 2025.

\bibitem[\protect\citeauthoryear{Schulman \bgroup \em et al.\egroup }{2017}]{PPO}
John Schulman, Filip Wolski, Prafulla Dhariwal, Alec Radford, and Oleg Klimov.
\newblock Proximal policy optimization algorithms.
\newblock {\em arXiv, 1707.06347}, 2017.

\bibitem[\protect\citeauthoryear{Sharma \bgroup \em et al.\egroup }{2019}]{DE-DDQN}
Mudita Sharma, Alexandros Komninos, Manuel L\'{o}pez-Ib\'{a}\~{n}ez, and Dimitar Kazakov.
\newblock Deep reinforcement learning based parameter control in differential evolution.
\newblock In {\em Proceedings of the Genetic and Evolutionary Computation Conference}, page 709–717, New York, NY, USA, 2019. Association for Computing Machinery.

\bibitem[\protect\citeauthoryear{Storn and Price}{1997}]{Storn_1997}
Rainer Storn and Kenneth Price.
\newblock Differential evolution – {A} simple and efficient heuristic for global optimization over continuous spaces.
\newblock {\em Journal of Global Optimization}, 11(4):341–359, 1997.

\bibitem[\protect\citeauthoryear{Tang and Yao}{2024}]{surveyL20}
Ke~Tang and Xin Yao.
\newblock Learn to optimize – {A} brief overview.
\newblock {\em National Science Review}, 11(8):nwae132, 04 2024.

\bibitem[\protect\citeauthoryear{Thrun}{1998}]{Thrun1998lifelong}
Sebastian Thrun.
\newblock {\em Lifelong Learning Algorithms}, pages 181--209.
\newblock Springer US, Boston, MA, 1998.

\bibitem[\protect\citeauthoryear{Tsaban \bgroup \em et al.\egroup }{2022}]{Tsaban_2022}
Tomer Tsaban, Julia~K. Varga, Orly Avraham, Ziv Ben-Aharon, Alisa Khramushin, and Ora Schueler-Furman.
\newblock Harnessing protein folding neural networks for peptide–protein docking.
\newblock {\em Nature Communications}, 13(1), January 2022.

\bibitem[\protect\citeauthoryear{TV \bgroup \em et al.\egroup }{2020}]{TV2019meta}
Vishnu TV, Pankaj Malhotra, Jyoti Narwariya, Lovekesh Vig, and Gautam Shroff.
\newblock Meta-learning for black-box optimization.
\newblock In {\em Machine Learning and Knowledge Discovery in Databases}, pages 366--381. Springer International Publishing, 2020.

\bibitem[\protect\citeauthoryear{Wang \bgroup \em et al.\egroup }{2020}]{wang2020incrementalTNNLS}
Zhi Wang, Han-Xiong Li, and Chunlin Chen.
\newblock Incremental reinforcement learning in continuous spaces via policy relaxation and importance weighting.
\newblock {\em IEEE Transactions on Neural Networks and Learning Systems}, 31(6):1870--1883, 2020.

\bibitem[\protect\citeauthoryear{Wang \bgroup \em et al.\egroup }{2023}]{BOsurvey}
Xilu Wang, Yaochu Jin, Sebastian Schmitt, and Markus Olhofer.
\newblock Recent advances in bayesian optimization.
\newblock {\em ACM Comput. Surv.}, 55(13s), July 2023.

\bibitem[\protect\citeauthoryear{Wang \bgroup \em et al.\egroup }{2024}]{surveyCL}
Liyuan Wang, Xingxing Zhang, Hang Su, and Jun Zhu.
\newblock A comprehensive survey of continual learning: Theory, method and application.
\newblock {\em IEEE Transactions on Pattern Analysis and Machine Intelligence}, 46(8):5362--5383, 2024.

\bibitem[\protect\citeauthoryear{Yang \bgroup \em et al.\egroup }{2025}]{yang2025graph}
Yifan Yang, Gang Chen, Hui Ma, Cong Zhang, Zhiguang Cao, and Mengjie Zhang.
\newblock Graph assisted offline-online deep reinforcement learning for dynamic workflow scheduling.
\newblock In {\em International Conference on Learning Representations}, 2025.

\bibitem[\protect\citeauthoryear{Yi \bgroup \em et al.\egroup }{2023}]{GSF-DQN}
Wenjie Yi, Rong Qu, Licheng Jiao, and Ben Niu.
\newblock Automated design of metaheuristics using reinforcement learning within a novel general search framework.
\newblock {\em IEEE Transactions on Evolutionary Computation}, 27(4):1072--1084, 2023.

\bibitem[\protect\citeauthoryear{Zhang \bgroup \em et al.\egroup }{2023}]{zhang2023catastrophic}
Tiantian Zhang, Xueqian Wang, Bin Liang, and Bo~Yuan.
\newblock Catastrophic interference in reinforcement learning: A solution based on context division and knowledge distillation.
\newblock {\em IEEE Transactions on Neural Networks and Learning Systems}, 34(12):9925--9939, 2023.

\bibitem[\protect\citeauthoryear{Zheng \bgroup \em et al.\egroup }{2022}]{zheng2022symbolic}
Wenqing Zheng, Tianlong Chen, Ting-Kuei Hu, and Zhangyang Wang.
\newblock Symbolic learning to optimize: Towards interpretability and scalability.
\newblock {\em arXiv, 2203.0657}, 2022.

\end{thebibliography}

\clearpage

\appendix
% \section{Solution Updating Rule Construction\label{ap:SEL}}

% We represent solution updating rules by tree-based symbolic equations and construct the tree with the LSTM model~\cite{Hochreiter_1997}, with the same method as used in SYMBOL~\cite{chen2024symbol}. The LSTM is trained by the PPO algorithm. The following subsections describe details in each part.

\section{State Representation\label{ap:state}}

The nine FLA metrics for state representation used in SYMBOL~\cite{chen2024symbol} are also used in this work (c.f., Section~\ref{sec:MDP}), including (i) the average distance between any pair of solutions in the current population, (ii) the average distance between each individual and the best individual in an optimization iteration, (iii) the average distance between each individual and the best-so-far solution, (iv) the average objective value gap between each individual and the best-so-far solution, (v) the average objective value gap between each individual and the best individual in an optimization iteration, (vi) the standard deviation of the objective values of the population in an optimization iteration, (vii) the potion of remaining optimization iterations, T denotes maximum optimization iterations for one run, (viii) the number of optimization iterations the optimizer stagnates improving, (ix) a binary value representing whether the optimizer finds a better solution than the best-so-far solution.

\section{Tree Structure Rule Representation\label{ap:LSTM_TreeBasedRule}}

We construct a solution updating rule (c.f., Section~\ref{sec:SEL_PPO}) as follows.
Following the design of SYMBOL, the operators of the rule, i.e., internal nodes of the tree, include addition ($+$), subtraction ($-$), and multiplication ($\times$). The operands of the rule, i.e., terminal nodes of the tree, include the manipulated incumbent solution ($x$), the best and worst solutions found so far ($x^*$ and $x^-$), the best solution for the $i$th solution, the differential vector (velocity) between consecutive steps of candidates ($\Delta x$), a randomly selected solution from the present population of solutions ($x_r$), and constant values ($c$). The height of a tree is set to be between $2$ and $5$, aligning with most human-designed optimizers.
Additionally, two distinct feed-forward layers are used to output the constant. The constant $c=\omega \times 10^\epsilon$ is represented in scientific notation,  distinct feed-forward layers are used to output the base $\omega$ and the exponent $\epsilon$ separately. The ranges are $\omega\in \{-1.0,-0.9,\dots,0.9,1\}$ and $\epsilon \in \{0,-1\}$. We use the identical vectorized tree embedding method as SYMBOL, to convert the current partially constructed tree into a vector, as a part of the LSTM's input.

\section{Lifelong Learning Details of LiBOG\label{ap:LiBOG}}

The EWC method for inter-task consolidation (c.f., Section~\ref{sec:inter}) involves the calculation of the importance of each parameter to the model on a task. The calculation of importance requires the gradient of each parameter under a set of experience trajectories. For good computation efficiency, we calculate the importance of the trajectories collected in the last epoch for model updating, instead of collecting new trajectories by running the final obtained model. Though the off-policy issue exists, due to that the model is updated in the last epoch after obtaining the experience trajectories, the experiment results show that LiBOG owns good performance. Algorithm~\ref{algo:LiBOG} demonstrates the detailed lifelong learning process of LiBOG.
\bc{To measure the difference in behaviors of the current model $\pi_{\theta}$ and the elite model $\pi_{\theta_e}$, EBC takes $D_{KL}(\pi_{\theta}(\cdot)\|\pi_{\theta_e}(\cdot))$ (c.f., Equation~\eqref{eq:ebc}) instead of commonly used $D_{KL}(\pi_{\theta_e}(\cdot)\|\pi_{\theta}(\cdot))$, as in literature it demonstrates better effectiveness in preventing $\theta$ from selecting unpromising actions that $\theta_e$ would take with very low (even zero) probability and stabilizing the learning, which is considered important in our case.}

\begin{algorithm}[htbp]
\caption{Lifelong learning of LiBOG\label{algo:LiBOG}}
\begin{algorithmic}[1]
\STATE \textbf{Input:} A sequence of tasks $(P_1,\dots,P_I)$, a guider optimizer $G$
\STATE \textbf{Output:} The model's final parameter values $\theta$
\STATE Initialize the parameters $\theta$ of the LSTM 
\STATE $\mathbf{\Omega} = \emptyset$
\STATE $\mathbf{\Theta^*} = \emptyset$
\FOR{$P_i \in \{P_1,\dots,P_I\}$}
\STATE $\theta_e \leftarrow \theta$
\STATE $\hat{R}_e \leftarrow -inf$
\FOR{$t \in \{1,\dots,N\}$}
\STATE $\mathbf{V} \leftarrow \emptyset$
\STATE Sample a set of $M$ problems $p$ from $P_i$
\STATE Experience trajectory set $\Psi \leftarrow \emptyset$
\FOR{$p_j \in p$}
\STATE $X \leftarrow$ initialize solutions 
\STATE $Y \leftarrow f_{p_i}(X)$
\STATE $v \leftarrow 0 $
\STATE $\tau \leftarrow \emptyset$
\WHILE {Optimization budget not run out}
\STATE $X'_G \leftarrow G(X)$
% \STATE $Y'_G \leftarrow f_{p_i}(X'_G)$
\STATE $O,\ell \leftarrow \pi_\theta(FLA(X))$
\STATE $X' \leftarrow O(X)$
\STATE $Y' \leftarrow f_{p_i}(X')$
\STATE $r \leftarrow R(X',X'_G,Y,Y')$
\STATE $v \leftarrow v+r$
\STATE $\tau \leftarrow \tau \cup (FLA(X),O,FLA(X'),r,\ell)$
\ENDWHILE
\STATE $\Psi \leftarrow \Psi \cup \tau$
\STATE $\mathbf{V} \leftarrow \mathbf{V} \cup v$
\ENDFOR
\IF{$\frac{1}{|\mathbf{V}|}\sum_{v\in \mathbf{V}} v \geq \hat{R}_e$}
\STATE $\hat{R}_e \leftarrow \frac{1}{|\mathbf{V}|}\sum_{v\in \mathbf{V}} v$
\STATE $\theta_e \leftarrow \theta$
\ENDIF
\STATE $\Delta \theta \leftarrow \eta \nabla_\theta (\mathcal{L}_{PPO}(\Psi) + \alpha \mathcal{CL}_{inter}(\theta,\mathbf{\Omega},\Theta^*) + \beta \mathcal{CL}_{intra}(\Psi,\theta_e))$
\STATE $\theta \leftarrow  \theta + \Delta \theta$
\ENDFOR
\STATE $\Omega_i \leftarrow$ calculate importance with $\Psi$
\STATE $\mathbf{\Omega} \leftarrow \mathbf{\Omega}\cup \Omega_i$
\STATE $\mathbf{\Theta^*} \leftarrow \mathbf{\Theta^*}\cup \theta$
\ENDFOR
\end{algorithmic}

\end{algorithm}

Besides, the EBC (c.f., Section~\ref{sec:intra}) needs to evaluate the performance of the newly obtained model to decide whether to update the elite model. To ensure computational efficiency, we evaluate the performance of the model obtained at the end of the previous epoch using experience trajectories sampled at the beginning of the current epoch, without introducing additional experience collection processes. For each epoch, we totally obtained $|\mathcal{J}|= N= 320$ trajectories for training.

\section{Experiment Details\label{ap:experiment}}

\subsection{Experiment Setting\label{ap:parameters}}

\begin{table*}[ht]
\centering
\caption{Summary of Functions.}
\begin{tabular}{c|c|p{5cm}|p{8cm}} % 使用 p{8cm} 控制列宽
\toprule
\textbf{Category} & \textbf{No.} & \textbf{Functions} & \textbf{Properties}    \\ \midrule
\multirow{1}{*}{\textbf{Uni-modal}} & 1 & Bent Cigar Function& Uni-modal, Non-separable, Smooth but narrow ridge   \\ \hline
\multirow{5}{*}{\textbf{Basic}}   & 2 & Schwefel's Function& Multi-modal, Non-separable, Local optima's number is huge\\ \cline{2-4}
  & 3 & bi-Rastrigin Function   & Multi-modal, Non-separable, Asymmetrical, Continuous everywhere yet differentiable nowhere \\ \cline{2-4}
  & 4 & Rosenbrock's plus Griewangk's Function   & Non-separable, Optimal point locates in flat area \\ \hline
\multirow{5}{*}{\textbf{Hybrid}}  & 5 & Hybrid Function 1 ($N = 3$)  & Multi-modal or Uni-modal, depending on the basic function\\ \cline{2-4}
  & 6 & Hybrid Function 2 ($N = 4$)  & Non-separable subcomponents\\ \cline{2-4}
  & 7 & Hybrid Function 3 ($N = 5$)  & Different properties for different variables subcomponents   \\ \hline
\multirow{3}{*}{\textbf{Composition}}   & 8 & Composition Function 1 ($N = 3$)  & Multi-modal\\ \cline{2-4}
  & 9 & Composition Function 2 ($N = 4$)  & Non-separable, Asymmetrical\\ \cline{2-4}
  & 10& Composition Function 3 ($N = 5$)  & Different properties around different local optima\\ 
  \bottomrule
\end{tabular}
\label{tab:functions}
\end{table*}
The detailed description of functions in the used dataset~\cite{cec2021so} is given in Table~\ref{tab:functions}.
To construct the lifelong learning MetaBBO scenario, We generate three task orders: (1) $P_0=P^U$, $P_1=P^B$, $P_2=P^H$, $P_3=P^C$. (2) $P_0=P^C$, $P_1=P^U$, $P_2=P^B$, $P_3=P^H$. (3) $P_0=P^U$, $P_1=P^C$, $P_2=P^H$, $P_3=P^B$. For LiBOG, \textit{restart} and \textit{fine-tuning} training, 320 problems are sampled for each task for training in each epoch. The optimization budget for optimizers to solve a problem once is 50000 function evaluations. The population size is set as 100. The learning rate is set to $\eta = 0.001$. For \textit{all-task} method, 32 problems are sampled for each function, forming a set of 320 training problems in each epoch. The models follow the same setting as in SYMBOL~\cite{chen2024symbol}. For the experiment for analysis of weights and ablation study, we use the first task order.

\subsection{Objective Value Normalization\label{ap:normalization}}
We normalize the objective value of a problem with the linear normalization method, as follows.
\begin{equation}
f^{norm} = \frac{f^{worst}-f}{f^{worst}-f^{opt}},
\end{equation}
where $f^{worst}$ is the largest (minimization problems) objective value found for all runs of all optimizers and the order on the function corresponding to the problem, $f^{opt}$ is the optimal objective, or the best found objective if the optimum is unknown, of the problem, and $f$ is the raw value before normalization. All results in the experiment are normalized using the above normalization method. Notably, the normalization, as well as the reward function (c.f., Equation~\eqref{eq:r}), do not strictly require a known optimal objective value. If the optimum is unknown, it can be easily approximated by the best found objective obtained from running an existing effective BBO optimizer.

\subsection{\bc{Runtime Efficiency} \label{ap:runtime}}
\bc{The two consolidation mechanisms in LiBOG introduce additional computational costs: (1) the cost in calculating $CL_{intra}$ and $CL_{inter}$, (2) the cost for computing $\Omega$.  
For training on all $M$ trajectories with trajectory length $\tau$ in the $i$th encountered task, where $|\theta|$ and $|\theta_c|$ represent the parameter sizes of the actor and critic networks, the time complexity of calculating $L_{PPO}$, $CL_{inter}$ and $CL_{intra}$ are $O((|\theta|+|\theta_c|) \cdot \tau \cdot M)$, $O(|\theta| \cdot i \cdot M)$ and $O(|\theta| \cdot \tau \cdot M)$, respectively. Considering cases where the number of tasks is much smaller than the number of iterations for solving a problem ($i \ll \tau$), the complexity of the overall loss function calculation is $O((|\theta|+|\theta_c|)\cdot \tau \cdot M)$, the same as $L_{PPO}$. The calculation of $\Omega$ is performed only once per task, with a complexity of $O(|\theta|)$, which is negligible. 
Empirically, the mean (± std.) training times for 10 runs of LiBOG and \textit{fine-tuning} are $124677.88(\pm30553.52)$ and $121794.56(\pm21144.62)$ seconds, respectively, indicating that the extra overhead for the consolidation mechanisms is insignificant.}

\subsection{Detailed Experiment Results\label{ap:results}}
\begin{figure}[htbp]
\centering
\includegraphics[width=0.9\linewidth]{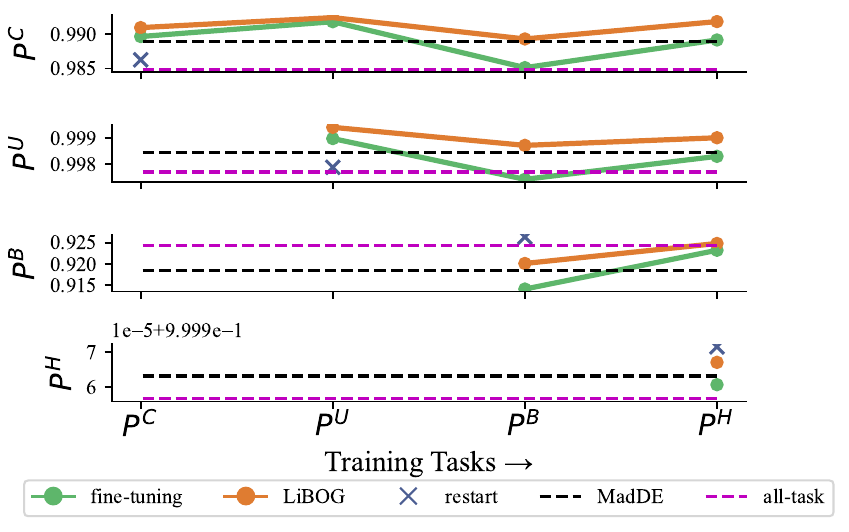}
\caption{Test performance on each learned task during the lifelong learning process of task order 1.}
\label{fig:forgetting_o1}
\end{figure}

\begin{figure}[htbp]
\centering
\includegraphics[width=0.9\linewidth]{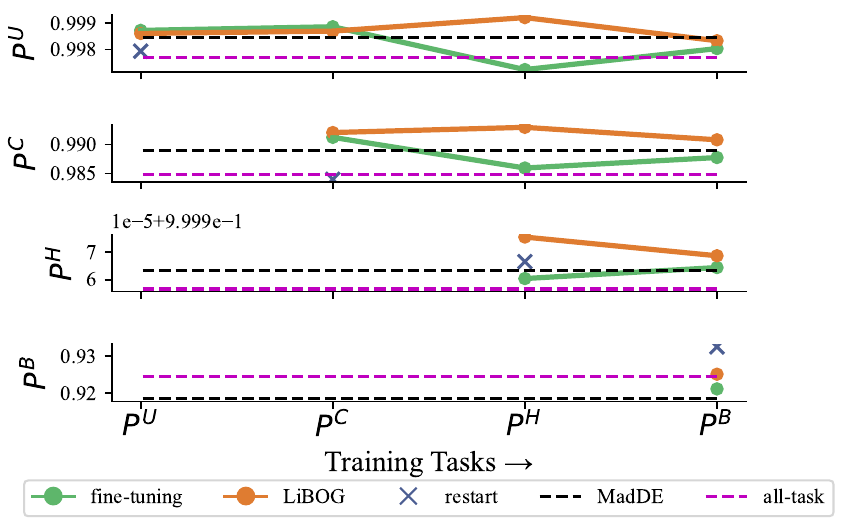}
\caption{Test performance on each learned task during the lifelong learning process of task order 2.}
\label{fig:forgetting_o2}
\end{figure}

\begin{figure}[htbp]
\centering
\includegraphics[width=0.9\linewidth]{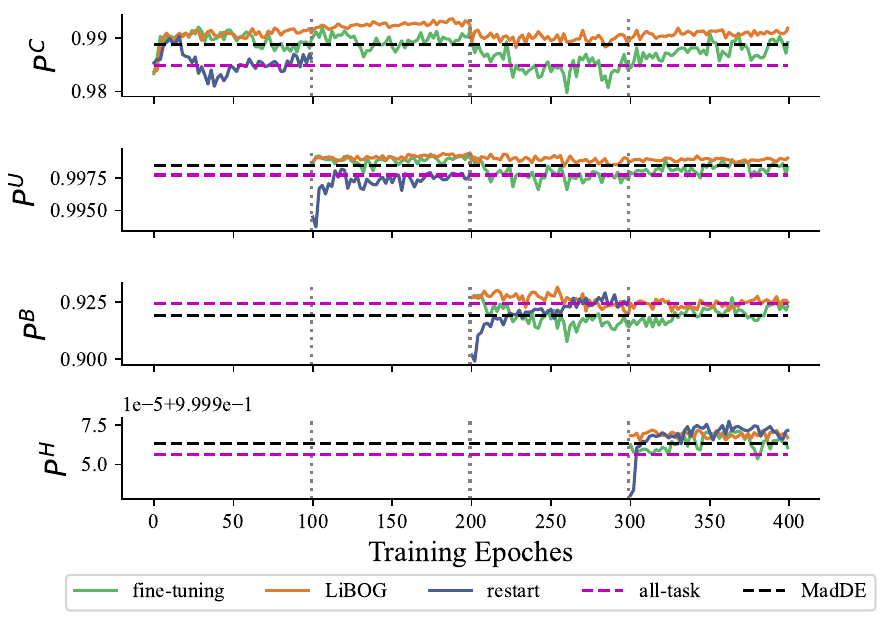}
\caption{Test performance of each model obtained during the lifelong learning process on each task of task order 1. }
\label{fig:training_curve_o1}
\end{figure}

\begin{figure}[htbp]
\centering
\includegraphics[width=0.9\linewidth]{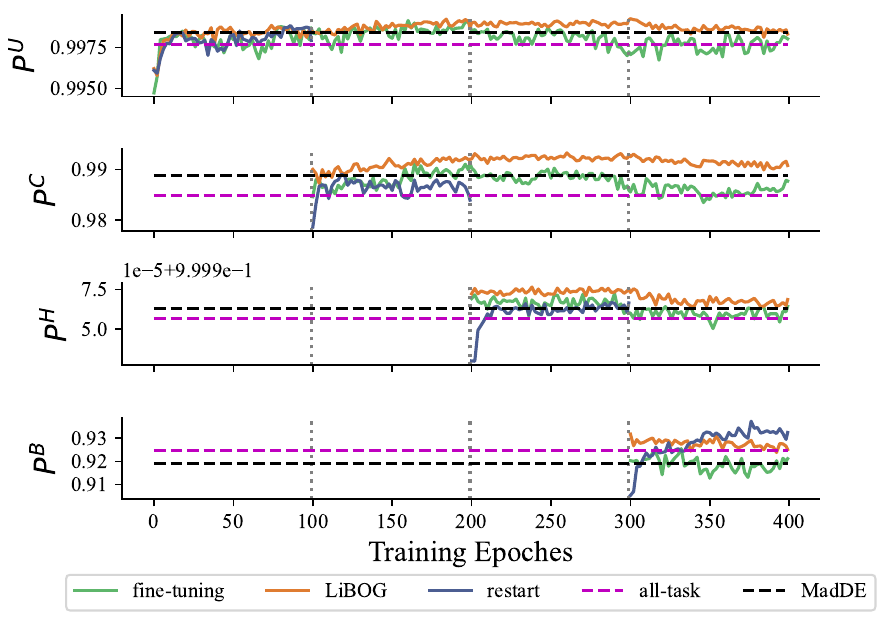}
\caption{Test performance of each model obtained during the lifelong learning process on each task of task order 2. }
\label{fig:training_curve_o2}
\end{figure}

The test performance on each task of the three task orders (c.f., Section~\ref{sec:effectiveness}) are listed in Table~\ref{tab:detail_result_o0},~\ref{tab:detail_result_o1} and~\ref{tab:detail_result_o2}.
The experiment results about forgetting (c.f., Section~\ref{sec:forgetting}) of the other order (order 1 and 2) are presented in Section~\ref{fig:forgetting_o1},~\ref{fig:forgetting_o2},~\ref{fig:training_curve_o1} and~\ref{fig:training_curve_o2}. 

\bc{Notably, although numbers in tables are very close to each other due to the normalization following SYMBOL, which takes the worst found objective value (several orders of magnitude larger than the obtained final solution's objective value of the problem) as the reference, the improvements are substantial. For instance, in the last task under task order $1$, the average raw objective values obtained by LiBOG and \textit{fine-tuning} are $2945.91$ and $4354.64$, respectively (with $0$ being the optimum).}

\begin{table*}[htbp]
\caption{The test performance on each task of task order 0.\label{tab:detail_result_o0}}
\begin{tabular}{c|c|c|c|c}
\toprule
\textbf{Method} & \textbf{$P_0=P^U$} & \textbf{$P_1=P^B$} & \textbf{$P_2=P^H$} & \textbf{$P_3=P^C$} \\
\midrule
\textbf{LiBOG}       & 0.999138(4.99549e-04) & 0.925060(4.58538e-03) & 0.999973(8.81920e-06)  & 0.992823(1.83466e-03) \\
\textbf{restart}     & 0.998082(1.71693e-03) & 0.921768(7.07112e-03) & 0.999971(0.995789e-06) & 0.991228(3.79684e-03) \\
\textbf{fine-tuning} & 0.998749(1.08664e-03) & 0.924385(8.71010e-03) & 0.999966(8.61156e-06)  & 0.989391(3.89779e-03) \\
\textbf{all-task}    & 0.997730(2.18734e-03) & 0.924816(1.30576e-02) & 0.999957(2.39998e-05)  & 0.984828(6.71194e-03) \\
\textbf{MadDE}       & 0.998373(2.14059e-04) & 0.920607(3.20594e-03) & 0.999962(2.12650e-06)  & 0.988859(8.60996e-04) \\
\bottomrule
\end{tabular}
\end{table*}

\begin{table*}[htbp]
\caption{The test performance on each task of task order 1.\label{tab:detail_result_o1}}
\begin{tabular}{c|c|c|c|c}
\toprule
\textbf{Method} & \textbf{$P_0=P^C$} & \textbf{$P_1=P^U$} & \textbf{$P_2=P^B$} & \textbf{$P_3=P^H$} \\
\midrule
\textbf{LiBOG}     & 0.99183701(1.7007303) & 0.99900601(5.9367504) & 0.92482501(4.7644403) & 0.99996701(8.6639206) \\
\textbf{restart}   & 0.98625601(5.1982303) & 0.99788801(1.4900003) & 0.92633901(1.1527802) & 0.99997101(9.6420106) \\
\textbf{fintuning} & 0.98915301(5.1037903) & 0.99830201(1.7454503) & 0.92321301(9.0005703) & 0.99996101(3.0681705) \\
\textbf{all-task}  & 0.98482801(6.7119403) & 0.99771001(2.2070103) & 0.92438401(1.3148102) & 0.99995701(2.3999805) \\
\textbf{MadDE}     & 0.98926901(1.1304103) & 0.99845701(2.6730304) & 0.91899801(2.9840503) & 0.99996401(1.7421006) \\
\bottomrule
\end{tabular}
\end{table*}

\begin{table*}[htbp]
\caption{The test performance on each task of task order 2.\label{tab:detail_result_o2}}
\begin{tabular}{c|c|c|c|c}
\toprule
\textbf{Method} & \textbf{$P_0=P^U$} & \textbf{$P_1=P^C$} & \textbf{$P_2=P^H$} & \textbf{$P_3=P^B$} \\
\midrule
\textbf{LiBOG}       & 0.998326(8.29036e-04) & 0.990810(1.47196e-03) & 0.999969(8.84050e-06) & 0.925086(4.96215e-03) \\
\textbf{restart}     & 0.997935(1.86867e-03) & 0.983968(6.29088e-03) & 0.999967(1.10313e-05) & 0.932434(1.37329e-02) \\
\textbf{fine-tuning} & 0.998032(1.60720e-03) & 0.987741(4.93295e-03) & 0.999964(1.12515e-05) & 0.921148(9.25448e-03) \\
\textbf{all-task}    & 0.997710(2.20701e-03) & 0.984828(6.71194e-03) & 0.999957(2.39998e-05) & 0.924513(1.31277e-02) \\
\textbf{MadDE}       & 0.998398(2.65031e-04) & 0.988903(9.53182e-04) & 0.999963(6.20836e-06) & 0.919315(3.43563e-03) \\ 
\bottomrule
\end{tabular}
\end{table*}

\begin{table*}[htbp]
\centering
\caption{Detailed results of weight setting analysis on each task .}
\begin{tabular}{c|c|c|c|c|c}
\toprule
\textbf{$\alpha$} & \textbf{$\beta$} & \textbf{$P_0=P^U$} & \textbf{$P_1=P^B$} & \textbf{$P_2=P^H$} & \textbf{$P_3=P^C$}\\
\midrule
\multirow{3}{*}{0.1}    & 0.1   & 0.999153(1.59e-04) & 0.930165(3.03e-03) & 0.999973(5.52e-06) & 0.993375(3.75e-04) \\
 & 1     & 0.999281(5.90e-04) & 0.927157(5.81e-03) & 0.999977(6.61e-06) & 0.991724(1.74e-03) \\
 & 10    & 0.999041(5.13e-04) & 0.936306(4.19e-03) & 0.999975(1.03e-05) & 0.993896(1.39e-03) \\ \hline
\multirow{3}{*}{1}      & 0.1   & 0.998915(4.81e-04) & 0.924905(4.68e-03) & 0.999968(9.01e-06) & 0.991552(2.63e-03) \\
 & 1     & 0.999138(5.00e-04) & 0.925060(4.59e-03) & 0.999973(8.82e-06) & 0.992823(1.83e-03) \\
 & 10    & 0.999013(3.83e-04) & 0.932025(5.48e-03) & 0.999969(5.31e-06) & 0.992057(1.66e-03) \\ \hline
\multirow{3}{*}{10}     & 0.1   & 0.999714(2.71e-04) & 0.935662(3.77e-03) & 0.999978(7.00e-06) & 0.993138(2.12e-03) \\
 & 1     & 0.999528(1.57e-04) & 0.935924(7.51e-03) & 0.999979(2.31e-06) & 0.994326(8.76e-04) \\
 & 10    & 0.998719(4.26e-04) & 0.929567(3.05e-03) & 0.999978(5.87e-06) & 0.993960(9.92e-04) \\ \hline
\multirow{4}{*}{Baselines} & {restart}     & 0.998082(1.72e-03) & 0.921768(7.07e-03) & 0.999971(9.96e-06) & 0.991228(3.80e-03) \\
 & {fine-tuning} & 0.998749(1.09e-03) & 0.924385(8.71e-03) & 0.999966(8.61e-06) & 0.989391(3.90e-03) \\
 & {all-task}    & 0.997730(2.19e-03) & 0.924816(1.31e-02) & 0.999957(2.40e-05) & 0.984828(6.71e-03) \\
 & {MadDE}       & 0.998446(3.37e-04) & 0.919432(2.20e-03) & 0.999963(2.52e-06) & 0.988834(8.49e-04)  \\
  \bottomrule
\end{tabular}
\label{tab:parameter_detail}
\end{table*}

\begin{table*}[htbp]
\centering
\caption{Detailed results of ablation study  on each task .}
\begin{tabular}{c|c|c|c|c}
\toprule
\textbf{Method}& \textbf{$P_0=P^U$} & \textbf{$P_1=P^B$} & \textbf{$P_2=P^H$} & \textbf{$P_3=P^C$}   \\
\midrule
\textbf{LiBOG}       & 0.999528(1.57e-04) & 0.935924(7.51e-03) & 0.999979(2.31e-06)  & 0.994326(8.76e-04) \\
\textbf{only-intra}  & 0.998879(1.98e-04) & 0.937388(5.25e-03) & 0.999976(4.94e-06)  & 0.992982(1.14e-03) \\
\textbf{only-inter}  & 0.997512(1.60e-03) & 0.918967(1.21e-02) & 0.999959(2.16e-05)  & 0.983777(6.33e-03) \\
\textbf{fine-tuning} & 0.998749(1.09e-03) & 0.924385(8.71e-03) & 0.999966(8.61e-06)  & 0.989391(3.90e-03) \\
\textbf{restart}     & 0.998082(1.72e-03) & 0.921768(7.07e-03) & 0.999971(0.996e-06) & 0.991228(3.80e-03) \\
\textbf{all-task}    & 0.997730(2.19e-03) & 0.924816(1.31e-02) & 0.999957(2.40e-05)  & 0.984828(6.71e-03) \\
\textbf{MadDE}       & 0.998446(3.37e-04) & 0.919432(2.20e-03) & 0.999963(2.52e-06)  & 0.988834(8.49e-04) \\
\bottomrule
\end{tabular}
\label{tab:ablation_detail}
\end{table*}

\end{document}